\documentclass[pdflatex,iicol,sn-basic]{sn-jnl}

\jyear{2022}%

\theoremstyle{thmstyleone}%
\theoremstyle{thmstyletwo}%

\theoremstyle{thmstylethree}%

\usepackage{hyperref}
\usepackage{color}
\usepackage{subfigure}
\usepackage{bigfoot}
\usepackage{framed}

\usepackage[normalem]{ulem}

\newcommand{\bs}{\boldsymbol}
\def\RR{ \mathbb R}
\newcommand{\refeq}[1]{Equation \eqref{#1}}

\newcommand{\ee}{\end{equation}}
\newcommand{\be}{\begin{equation}}
\newcommand{\ec}{\end{center}}
\newcommand{\bc}{\begin{center}}
\newcommand{\eea}{\end{eqnarray}}
\newcommand{\bea}{\begin{eqnarray}}
\newcommand{\bd}{\begin{description}}
\newcommand{\ed}{\end{description}}
\newcommand{\bi}{\begin{itemize}}
\newcommand{\ei}{\end{itemize}}
\newcommand{\pa}{\partial}

\newcommand{\bt}{\bs{\theta}}

\begin{document}

\title[Semi-supervised Invertible Neural Operators for Bayesian Inverse Problems]{Semi-supervised Invertible Neural Operators for Bayesian Inverse Problems}


\author[1]{\fnm{Sebastian} \sur{Kaltenbach}}\email{sebastian.kaltenbach@tum.de}

\author[2]{\fnm{Paris} \sur{Perdikaris}}\email{pgp@seas.upenn.edu}

\author*[1,3]{\fnm{Phaedon-Stelios} \sur{Koutsourelakis}}\email{p.s.koutsourelakis@tum.de}

\affil[1]{\orgdiv{Professorship of Data-driven Materials Modeling}, \orgname{School of Engineering and Design, Technical University of Munich }, \orgaddress{\street{Boltzmannstr. 15}, \city{Garching}, \postcode{85748}, \country{Germany}}}

\affil[2]{\orgdiv{Department of Mechanical Engineering and Applied Mechanics}, \orgname{University of Pennsylvania}, \orgaddress{\city{Philadelphia}, \postcode{19104}, \country{USA}}}

\affil[3]{ \orgname{Munich Data Science Institute (MDSI - Core member)}, \orgaddress{ \city{www.mdsi.tum.de}}}


\abstract{Neural Operators offer a powerful, data-driven  tool for  solving parametric PDEs as they can represent maps between infinite-dimensional function spaces. In this work, we employ physics-informed Neural Operators in the context of high-dimensional, Bayesian inverse problems.  Traditional solution strategies necessitate an enormous, and frequently infeasible, number of forward model solves, as well as the computation of parametric derivatives. In order to enable efficient solutions, we extend Deep Operator Networks (DeepONets) by employing a RealNVP architecture which yields an invertible and differentiable  map between the parametric input and the branch-net output. This allows us to construct accurate approximations of the full posterior,  irrespective of the number of observations and the magnitude of the observation noise,  without any need for  additional forward solves  nor  for cumbersome, iterative sampling procedures. We demonstrate the efficacy and accuracy of the proposed methodology in the context of inverse problems for three benchmarks:  an anti-derivative equation, reaction-diffusion dynamics and flow through porous media.} 




\maketitle

\section{Introduction}

Nonlinear Partial Differential Equations (PDEs) depending on high- or even infinite-dimensional parametric inputs are ubiquitous in applied physics and engineering and appear in the context of several problems such as model calibration and validation or model-based design/optimization/control.
In all these cases, they must be solved repeatedly  for different values of the input parameters which poses an often insurmountable obstacle  as each of these simulations can  imply a significant computational cost.  
An obvious way to overcome these difficulties is to develop less-expensive but accurate  surrogates which can be used on their own or in combination with a reduced number of runs of the high-fidelity, expensive, reference solver. The construction of such surrogates has been based on physical/mathematical considerations or data i.e. input-output pairs (and sometimes derivatives). Our contribution belongs to the latter category of data-driven surrogates which  has attracted a lot of attention in recent years  due to the significant progress in the fields of statistical or machine learning  \citep{koutsourelakis2016big,karniadakis_physics-informed_2021}. We emphasize however that unlike typical supervised learning problems in data sciences, in the context of computational physics there are several distinguishing features.  Firstly, surrogate construction is by definition a Small (or smallest possible) Data problem. The reason we want to have a surrogate in the first place is to avoid using the reference solver which is the one that generates the training data. Secondly, pertinent problems are rich in domain knowledge which should be incorporated as much as possible, not only in order to reduce the requisite training data but also to achieve higher predictive accuracy particularly in out-of-distribution settings. In the context of Bayesian inverse problems which we investigate in this paper, one does not know a priori where the posterior might be concentrated in the parametric space and  cannot guarantee that all such regions will be sufficiently represented in the training dataset. Nevertheless the surrogate learned must be accurate enough in these regions in order to resolve the sought posterior.\\

Data-driven surrogates which are trained in an offline phase and are subsequently used for various downstream tasks have attracted a lot of attention in recent years \citep{bhattacharya2020model}. Most of these surrogates are constructed by learning a non-linear operator, e.g. a mapping between function spaces and thus between the inputs and the outputs of the PDE, which may depend on additional input parameters. A notable such strategy 
 based on Deep Learning are the   Physics-informed Neural Networks (PINNs) \citep{lagaris1998artificial,raissi2019physics}.  An alternative is offered by Deep Operator Networks (DeepONets, \citep{lu2021learning, wang2021learning}), which in contrast to PINNs, not only take  the spatial and temporal location as an input but can also account for the dependence of the PDE solution on input parameters  such as the viscosity in the Navier-Stokes equation. Furthermore, Fourier Neural Networks \citep{li2020fourier} have shown promising results by parametrizing the integral kernel directly in Fourier Space and thus restricting the operator to a convolution. Finally,  the Learning Operators with Coupled Attention (LOCA) framework \citep{kissas2022learning} builds upon the well-known attention mechanism that has already shown promising results in natural language processing.\\
We note that all of the Deep Learning frameworks mentioned fulfill the universal approximation theorem and, under certain conditions, can  approximate the non-linear operator to arbitrary accuracy. Another option, is offered by the  Optimizing a Discrete Loss (ODIL, \cite{karnakov2022optimizing}) framework. It   does not rely on  Deep Learning  and was shown to be faster than PINNs due to the reduced number of tunable parameters  but  can only approximate the solution on a discrete grid.\\
Apart from the aforementioned techniques and for time-dependent PDEs in particular, the solution can be  approximated by  methods based on Koopman-operator theory \citep{koopman_hamiltonian_1931} which identifies a transformation of the original system that gives rise to  linear dynamics \citep{klus_data-driven_2018}. Nevertheless, these methods  \citep{lee2020model,gin2019deep,champion2019discovery} usually require a  large set of reduced-order coordinates  or an effective encoder/decoder structure. Especially for physical systems, the  restricted dynamics can be endowed with stability and physical, inductive bias \citep{kaltenbach2021physics, kalia2021learning,kaltenbach2020incorporating}.\\

A common limitation of the aforementioned  architectures  is that they   usually learn only the forward  operator whereas for the solution of an inverse problem, its inverse would be more useful. In this work, we extend the DeepONet framework by replacing parts of  the previously proposed  neural-network architecture with an invertible one. To the authors' best knowledge, we are thus presenting the first invertible Neural Operator framework. This allows one to perform both forward and inverse passes with the same neural network and the forward and inverse operators can be learned simultaneously. In particular, we make use of the RealNVP architecture \citep{dinh2016density} which has an analytical inverse.\\
Furthermore we make use of both labeled and unlabeled (i.e. only inputs and residuals) training data  in a physics-aware, semi-supervised approach. While the use of labeled training data is straight-forward, unlabeled training data are incorporated by using the governing equations 
 and minimizing the associated residuals, similarly to the physics-informed DeepONet \citep{wang2021learning}. Since it is easier and less-expensive to procure  unlabeled data in comparison to labeled ones, this  leads to significant efficiency gains. Even though our algorithm  can produce accurate  predictions without any labeled training data and by  using only a physics-informed loss, we observe empirically that the addition of labeled training data generally  improves the results.\\
Finally, we show that the proposed invertible DeepONet can be used to very efficiently solve Bayesian inverse problems, i.e. to approximate the whole posterior distribution,  without any need for multiple likelihood evaluations and cumbersome iterations as required by alternative inference schemes such as Markov Chain Monte Carlo  (MCMC, \cite{beskos_geometric_2017}) or Sequential Monte Carlo (SMC, \cite{koutsourelakis_multi-resolution_2009}) or Stochastic Variational Inference (SVI, \cite{detommaso_stein_2018}). In particular, we propose a novel approximation that employs a mixture of Gaussians, the parameters of which are computed semi-analytically. When the proposed Neural Operator framework is trained solely on unlabeled data, this means that we can obtain the solution to the (forward and) inverse problem without ever solving the underlying PDE.
While Deep Learning has been successfully applied to inverse problems before \citep{adler2017solving, ardizzone2018analyzing,mo_deep_2019}, our work differs by making use of a fully invertible, operator-learning architecture which leads to highly efficient approximation of the whole posterior. 

The rest of the  paper is structured as follows. In section \ref{sec:method} we review the basic elements of invertible neural networks (NNs)  and DeepoNets and subsequently illustrate how these can be combined and trained with labeled and unlabeled data. Furthermore we present how the resulting invertible DeepONet can be employed in order to approximate the posterior of a model-based, Bayesian inverse problem at minimal additional cost. We illustrate several features of the proposed methodology and assess its performance in section \ref{sec:num} where it is applied to  a reaction-diffusion PDE and a  Darcy-diffusion problem.  The cost and accuracy of the posterior approximation in the context of pertinent Bayesian inverse problems are demonstrated in section \ref{sec:bayes}.
Finally, we conclude in section \ref{sec:con} with a summary of the main findings and a discussion on the (dis)advantages of the proposed architecture and potential avenues for   improvements. 

\section{Methodology}
\label{sec:method}
We first review some basic concepts of invertible neural networks and  DeepONets. We subsequently present  our novel contributions which consist of an  invertible DeepONet  architecture and  its use for solving efficiently Bayesian inverse problems.

\subsection{Invertible Neural Networks}
\label{sec:invnn}
Neural Networks are in general not invertible which restricts their application in  problems requiring inverse operations. Invertibility can be achieved by adding a momentum term \citep{sander2021momentum}, restricting the Lipschitz-constant of each layer to be smaller than one \citep{behrmann2019invertible} or using special building blocks \citep{dinh2016density}. These formulations have primarily been developed for flow-based architectures but we will apply them to operator learning within this work.\\
In particular, we make use of the  RealNVP \citep{dinh2016density} as this architecture enables an analytical inverse which ensures  efficient computations. Each RealNVP building block consists of the transformation below which includes two neural networks denoted by  $k(.)$ and $r(.)$. Given a $D$ dimensional input $\bs{x}=\{x_i\}_{i=1}^D$ of an invertible layer, the output $\bs{y}=\{y_i\}_{i=1}^D$ is obtained as follows:
\begin{equation}
    y_{1:d}=x_{1:d}
\end{equation}
\begin{equation}
    y_{d+1:D}=x_{d+1:D} \circ exp(k(x_{1:d})) + r(x_{1:d}),
    \label{eq:2}
\end{equation}
where $d < D$.
Here, $\circ$ is the Hadamard or element-wise product and $d$ is usually chosen to be half of the dimension of the input vector i.e. $d=D/2$.\\
As only $d$  of the components are updated,  the input entries after each building block are permuted, e.g. by reversing the vector,  to ensure that after a second building block all of them are modified. Therefore, for $d=D/2$, at least two building blocks are needed in order to modify all entries. We note, that the dimension of the input cannot change and it needs to be identical to the dimension of the output. The two neural networks involved can consist of arbitrary layers as long as their output and input dimensions are consistent with \refeq{eq:2}.\\
The maps defined can be easily inverted which leads to the following equations:
\begin{equation}
    x_{1:d}=y_{1:d}
\end{equation}
\begin{equation}
    x_{d+1:D}=(y_{d+1:D}-r(x_{1:d})) \circ exp(-k(x_{1:d}))
\end{equation}
We note that due to this structure, the Jacobian is lower-triangular  and its determinant can be obtained by multiplying the diagonal entries only.

\subsection{DeepONets}
\begin{figure*}[h]
 \begin{minipage}[c]{0.4\linewidth}
 \centering
    \includegraphics[trim=80 620 120  50, clip,scale=0.55]{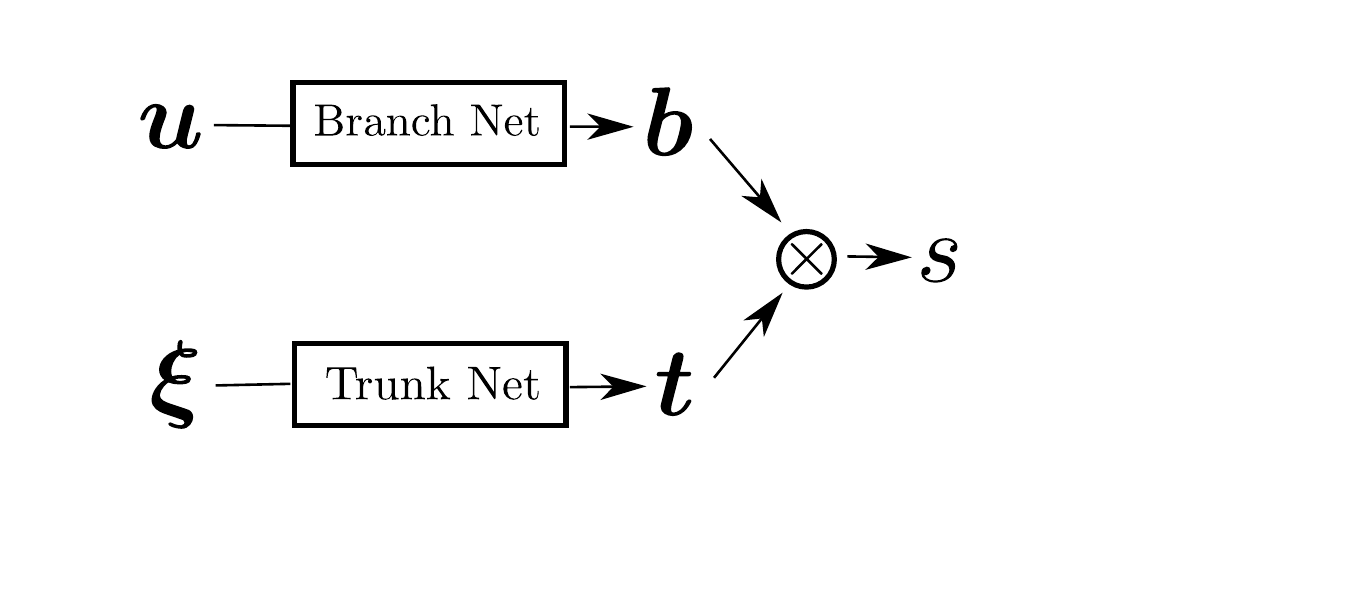}
       \end{minipage}
 \hfill
 \begin{minipage}[c]{0.4\linewidth}
  \centering
    \includegraphics[trim=40 620 120  50, clip,scale=0.45]{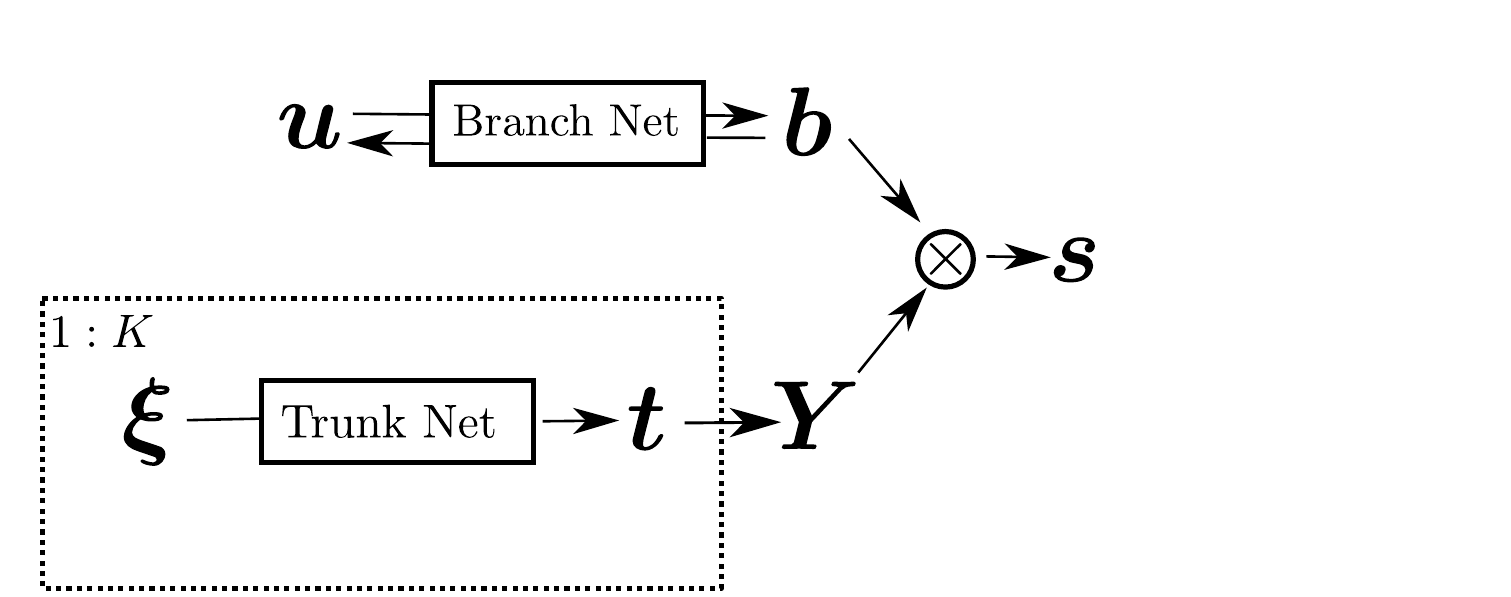}
    \label{fig:DeepONet}
     \end{minipage}
\caption{(Left) Classical DeepONet \citep{lu2019deeponet} and (Right) proposed Invertible DeepONet architecture}  
    \label{fig:Inv_DeepONet}
\end{figure*}

Before presenting our novel architecture for invertible DeepONets, we briefly  review the original DeepONet formulation by \cite{lu2021learning}. DeepONets have been developed to solve parametric PDEs and significantly extend  the Physics-Informed Neural Network (PINNs, \cite{raissi2019physics}) framework  as no additional training phase is required if the input parameters of the PDE are changed. 
We consider a,  potentially nonlinear and time-dependent, PDE with an input function $u \in \mathcal{U}$ and solution function $s \in \mathcal{S}$ where $\mathcal{U,S} $ are appropriate Banach spaces. The former can represent e.g. source terms, boundary or initial conditions, material properties. Let:
\begin{equation}
    \mathcal{N}(u,~s)(\bs{\xi})=0
    \label{eq:PDE}
\end{equation}
denote the governing PDE where $\mathcal{N}:\mathcal{U}\times \mathcal{S} \to \mathcal{V}$ is an appropriate differential operator and $\bs{\xi}$ the spatio-temporal coordinates. Furthermore, let:
\be
\mathcal{B}(u,~s)(\bs{\xi})=0
    \label{eq:BC}
\ee
denote the operator $\mathcal{B}:\mathcal{U}\times \mathcal{S} \to \mathcal{V}$ associated with the boundary or initial conditions. Assuming that the solution $s$ for each $u \in \mathcal{U}$ is unique, we denote with   
$\mathcal{G}: \mathcal{U}  \to \mathcal{S}$  the solution operator that maps from any input ${u}$ to the corresponding solution ${s}$. The goal of DeepONets is to approximate it with an operator  $G_{\bt}$ that depends on tunable parameters $\bt$.
The latter can yield an approximation to the actual solution at any spatio-temporal point $\bs{\xi}$ which we denote by $G_{\bt}(\bs{\xi})$.  It is based on a separated representation \citep{lu2021learning}\footnote{We omit  the NN parameters $\bt$ on the right-hand side in order to simplify the notation.}: 
\begin{equation}
    G_{\bt}(u)(\boldsymbol{\xi})=\sum_{j=1}^Q b_{j}\left(~\underbrace{u(\boldsymbol{\eta}_1),...,u(\boldsymbol{\eta}_F)}_{\bs{u}} \right)~t_{j}(\boldsymbol{\xi})
    \label{eq:donet}
\end{equation}
and consists of the so-called {\em branch network} whose terms $b_{j}$ depend on the values of the input function $u$ at $F$ fixed spatio-temporal locations\footnote{These points are usually chosen to be uniformly distributed over the entire domain, but it is also possible to increase their density in  certain areas, e.g. with  high variability.}   $\{\bs{\eta}_l\}_{l=1}^F$ which we summarily denote with the vector $\bs{u} \in \RR^F$, and the so-called {\em trunk network} whose terms $t_{j}$  depend on the spatio-temporal coordinates $\bs{\xi}$ (see Figure \ref{fig:DeepONet}).  Both networks have trainable weight and bias parameters which we denote collectively by $\bt$.
We emphasize that, once trained, the DeepONet can provide predictions of the solution at {\em any spatio-temporal location $\bs{\xi}$}, a feature that is very convenient in the context of inverse problems as the same DeepONet can be used for solving problems with  different sets of observations.   \\

We note that in the next section, we will use a vectorized formulation of  \refeq{eq:donet} and process various spatio-temporal coordinate datapoints together as this is needed to ensure invertibility of the DeepONet.
Labeled data can be used for training which consist of pairs of ${u}$ and corresponding solutions ${s}=\mathcal{G}({u})$ evaluated at certain spatio-temporal locations.
Unlabeled training data (i.e. only inputs) can also be employed in a physics-informed approach as introduced by \cite{wang2021learning}, by including the governing PDE in \refeq{eq:PDE} in an additional loss term as discussed section \ref{sec:loss}.

\subsection{Invertible DeepONets}
\label{sec:idon}

The invertible RealNVP introduced in section \ref{sec:invnn} is employed exclusively on the branch network i.e. we assume that:
\be
D=F=Q
\ee
and the input $\bs{x}$ of section \ref{sec:invnn} is the vector $\bs{u} \in \RR^F$ containing the values of the PDE-input at $D=F$ spatio-temporal locations whereas the output $\bs{y}$ of section \ref{sec:invnn}  is now the $D=Q$ values of the branch net $\bs{b}=[b_1,\ldots,b_Q]^T  \in \RR^D$. We note that this restriction regarding the equality of the dimension of the input $\bs{u}$ and the output of the branch network $\bs{b}$ is due to the use of an invertible architecture. 
As  a consequence, the dimension of the trunk-network output i.e. $\{t_j(\bs{\xi})\}_{j=1}^Q$ is also the same as the dimension of $\bs{u}$. This requirement does not reduce the generality of the methodology advocated as $Q$ is a free parameter in the definition of the operator $G_{\bt}$ in \refeq{eq:donet}.

In view of the inverse problems we would like to address, we consider   $K$ spatio-temporal locations, $\{ \bs{\xi}_k \}_{k=1}^K$ and we denote  with $\bs{s} \in \RR^K$ the vector containing  the PDE-solution's values at these locations i.e. $\bs{s}=\left[ s(\bs{\xi}_1), \ldots, s(\bs{\xi}_K)\right]^T$.  Finally we denote with $\bs{Y}$ the $K \times D$ matrix constructed by the values of the trunk network outputs at the aforementioned locations, i.e.:
\be
    \boldsymbol{Y}=\begin{bmatrix} t_1(\bs{\xi}_1)&  ... &t_D(\bs{\xi}_1) \\...& &...\\ t_1(\bs{\xi}_K) &...&t_D(\bs{\xi}_K) \end{bmatrix}.
    \label{eq:y}
\ee
As a result of \refeq{eq:donet}, we can write that:
\be
 \boldsymbol{s}=\boldsymbol{Y} \boldsymbol{b} 
 \label{eq:svbv}
\end{equation}
As the matrix $\bs{Y}$ is in general non-invertible, one can determine $\bs{b}$ given $\bs{s}$ by solving a least-squares problem,  i.e.:
\be
\min_{\bs{b}} ~\| \bs{s} -\bs{Y} \bs{b} \|^2_2
\label{eq:min1}
\ee
or  a better-behaved, regularized version thereof:
\be
\min_{\bs{b}} ~\| \bs{s} -\bs{Y} \bs{b} \|^2_2 + \epsilon \| \bs{b} \|^2_2
\label{eq:min2}
\ee
where a small value is generally sufficient for the regularization parameter $\epsilon <<1$. We note that given $\bs{s}$ and once $\bs{b}$ has been determined by solving \refeq{eq:min1} or \refeq{eq:min2}, we can make  use  of  the invertibility of the branch net in order to obtain the input vector $\bs{u}$. While  other approaches are possible in order to determine $\bs{b}$, we recommend using the regularized, least-squares formulation,  as this led to robust results in our experiments. It is nevertheless important to use the same method  during training and when  deterministic  predictions are sought, since different methods can lead to different  $\bs{b}$'s for the same $\bs{s}$. We note that in the proposed method for the solution of Bayesian inverse problems (see Section \ref{sec:bip}), no use of \refeq{eq:min2} is made  except for the training of the DeepONet (see Section \ref{sec:loss}).

For the ensuing equations we denote the forward map implied by \refeq{eq:svbv} as:
\be
\bs{s}=\mathcal{F}_{\bt}(\boldsymbol{u},\boldsymbol{Y})
\label{eq:fmap}
\ee
and the inverse obtained by the two steps described above as:
\be
\bs{u}=\mathcal{I}_{\bt}(\boldsymbol{s},\boldsymbol{Y})
\label{eq:imap}
\ee
where we explicitly account for the NN parameters $\bt$. 
 
 \subsection{A Semi-supervised Approach for Invertible DeepONets}
 \label{sec:loss}
 
As  mentioned earlier and in order to train the invertible DeepONet proposed, i.e.  to find the optimal values for the parameters $\bt$, we employ both labeled (i.e pairs of PDE-inputs $u$ and PDE-outputs $s$) and unlabled data (i.e. only PDE-inputs $u$) in combination with the governing equations.
The loss function $L$ employed is therefore decomposed into two parts as\footnote{All loss functions depend on $\bt$ which we omit in order to simplify the notation.}:
 \begin{equation}
    L = L_{labeled} + L_{unlabeled}
\end{equation}
 The first term $L_{labeled}$ pertains to  the labeled data  and is further decomposed as:
 \begin{equation}
     L_{labeled}=L_{l,forward}+L_{l,inverse}
 \end{equation}
 
 Without loss of generality and in order to keep the notation as simple as possible we assume that $N_l$ pairs of labeled data are available,  each of which consists of the values of the PDE-input $u$ at $D$ locations which we denote with $\bs{u}^{(i)} \in \RR^D, ~i=1, \ldots N_l$ and the values of the PDE-output at $K$ spatio-temporal locations which we denote with $\bs{s}^{(i)} \in \RR^K,~i=1, \ldots N_l$. If the $K\times D$ matrix $\bs{Y}$ is defined as in \refeq{eq:y}
  and in view of the forward (\refeq{eq:fmap}) and inverse (\refeq{eq:imap}) maps defined earlier, we  write:
 \be
 L_{l,forward}=\frac{1}{N_l}\sum_{i=1}^{N_l} \|\bs{s}^{(i)}-\mathcal{F}_{\bt}(\bs{u}^{(i)},\bs{Y}) \|^2_2
 \label{eq:lforw}
 \ee
 and:
 \be
 L_{l,inverse}=\frac{1}{N_l} \sum_{i=1}^{N_l} \| \bs{u}^{(i)}-\mathcal{I}_{\bt}(\boldsymbol{s}^{(i)},\boldsymbol{Y} ) \|^2_2.
 \label{eq:lback}
 \ee
 By employing both loss terms, the NN parameters $\bt$ can balance the accuracy of the approximation in both maps.
 
 Furthermore and assuming $N_u$ PDE-inputs are available each of which is evaluated at $D$ spatio-temporal points $\{ \bs{\xi}^{(l)} \}_{l=1}^D$  with $\bs{u}^{(i)} \in \RR^D$ denoting these values, we express the  $L_{unlabeled}$ loss term  as:
\be 
L_{unlabeled}=L_{BC}+L_{res}+L_{u,inverse}.
\ee
The first $L_{BC}$ and second $L_{res}$ terms are physics-informed \cite{wang2021learning} and account for the residuals in the boundary (and/or initial)  conditions and the governing  PDE respectively. In the case of $L_{BC}$ we select $N_B$ (uniformly distributed) points along the boundary, say $\bs{\xi}_B^{(j)}, l=1,\ldots, N_B$. 

Then, in view of \refeq{eq:BC}, we employ:
 \be
 L_{BC}=\frac{1}{N_u N_B}\sum_{i=1}^{N_u} \sum_{l=1}^{N_B}
 \| 
 \mathcal{B} ( u^{(i)}, {G}_{\bt}(u^{(i)}) )(\bs{\xi}_B^{(l)} ) 
 \|^2_2
 \label{eq:ubc}
 \ee
 In the interior of the problem domain and in view of \refeq{eq:PDE}, we employ a loss: 
\be
 L_{res}= \frac{1}{N_u~N_{res}} \sum_{i=1}^{N_u} \sum_{l=1}^{N_{res}} \| \mathcal{N}( u^{(i)},  {G}_{\bt}({u}^{(i)})(\bs{\xi}^{(l)} ) \|^2_2
 \label{eq:lres}
 \ee
 which involves $N_{res}$ collocation points.
 
 The third term $L_{u,inverse}$ pertains to the forward and inverse maps in Equations (\ref{eq:fmap}), (\ref{eq:imap}) and can be expressed as:
 \be
  L_{u,inverse}= \frac{1}{N_u} \sum_{i=1}^{N_u} \| \bs{u}^{i}-\mathcal{I}_{\bt}(\mathcal{F}_{\bt}(\bs{u}^{(i)} ,\boldsymbol{Y} ),\bs{Y} )) \|^2_2
  \label{eq:uinv}
 \end{equation}
where the matrix $\bs{Y}$ is defined as in \refeq{eq:y}.

The minimization of the combined loss $L$, with respect to the NN parameters $\bt$   of the branch and trunk network, is performed with  stochastic gradient descent and the  ADAM \citep{kingma2014adam} scheme in particular. Gradients of the loss were computed using the automatic differentiation tools of the JAX  library \citep{jax2018github}.
We finally note that the $D$ spatioemporal locations need not be the same nor do they need to be equal in number in all data instances as assumed in the equations above. In such cases the vector of the observables and the  matrices $\bs{Y}$  involved would differ which would further complicate the notation but the same DeepONet parameters $\bt$ would appear in all terms.

\subsection{Invertible DeepONets for Bayesian inverse problems}
\label{sec:bip}

In this section we discuss how the invertible DeepONets proposed and trained as previously discussed, can be used to efficiently  approximate the solution of a Bayesian inverse problem in the presence of, potentially noisy, observations as well as prior uncertainty about the unknowns.
A central role is played by the readily available invertible map which the RealNVP architecture  affords. In particular, let $\hat{\bs{s}} \in \RR^K$ denote a vector of noisy observations of the PDE-solution at certain $K$ spatio-temporal locations. These are assumed to be related to the PDE-solution's values at these locations, denoted summarily by  $\bs{s} \in \RR^K$,  as follows:
\be
\hat{\bs{s}}=\bs{s}+\sigma~\bs{\eta}, \qquad \bs{\eta}\sim\mathcal{N}(\bs{0,I}).
\label{eq:BIP_noise}
\ee
where $\sigma^2$ is the  variance of the observational noise.
This in turn defines a conditional density (likelihood) $p(\hat{\bs{s}} \mid  \bs{s})$:
\be 
p(\hat{\boldsymbol{s}} \mid  \bs{s})= \mathcal{N}(\hat{\bs{s}} \mid \boldsymbol{s}, \sigma^2\bs{I}).
\ee

In the context of a Bayesian formulation  and given the implicit dependence of the PDE-output $\bs{s}$ on $\bs{u}$, the likelihood would be combined with the a prior density $p_u(\bs{u})$ on the PDE-inputs in order to define the sought posterior:
\be
p(\bs{u}   \mid   \hat{\bs{s}} ) \propto p(\hat{\bs{s}} \mid  \bs{s})~p_u(\bs{u}).
\nonumber
\ee
Even if the trained DeepONet were used to infer $p(\bs{u}   \mid   \hat{\bs{s}} )$ (e.g. using MCMC) several evaluations would be needed especially if the dimension of $\bs{u}$ was high.
In the sequel we demonstrate how one can take advantage of the invertible NN architecture in order to obtain a semi-analytic approximation of the posterior in the form of a mixture of Gaussians and by avoiding iterative algorithms like MCMC altogether.

We  note first that by combining the likelihood with \refeq{eq:svbv}, we can write it in terms of  the $D-$dimensional, branch-network output vector $\bs{b}$ as:
\be
p(\hat{\boldsymbol{s}} \mid \bs{b})= \mathcal{N}(\hat{\bs{s}} \mid \bs{Y~b}, \sigma^2\bs{I}).
\label{eq:likeb}
\ee
Since  $\bs{u} \in \RR^D$  is related to $\bs{b}$ through the invertible RealNVP $\bs{b}_{NN}: \RR^D \to \RR^D$, we can also obtain a  prior density $p_b(\bs{b})$ on $\bs{b}$ as:
\be
p_b(\bs{b})= p_u(\bs{b}_{NN}^{-1}(\bs{b}) )~J(\bs{b})
\ee
where 
 $\bs{b}_{NN}^{-1}$ denotes the  inverse and  $J(\bs{b}) =\lvert \frac{\pa \bs{b}_{NN}^{-1}}{\pa \bs{b}} \rvert $ is the determinant of its Jacobian. The latter, as mentioned in section \ref{sec:invnn}, is a triangular matrix and its determinant can be readily computed at a cost $\mathcal{O}(D)$.

We  choose not to directly operate with the prior $p_b(\bs{b})$, but construct an approximation $p_{b,G}(\bs{b})$ to this in the form of a mixture of  $D-$dimensional Gaussians  as this allows as to facilitate subsequent steps in finding the  posterior. In particular:
 \be
 p_{b,G}(\bs{b})=\sum_{m=1}^M w_j~\mathcal{N}(\bs{b} \mid \bs{m}_{b,m}, \bs{S}_{b,m})
 \label{eq:mixprior}
\ee
where $M$ denotes the number of mixture components and $\bs{m}_{b,m}$,  $\bs{S}_{b,m}$ the mean vector and covariance matrix of the $m^{th}$ component respectively. Such an approximation can be readily computed, e.g. using Variational Inference \citep{wainwright_graphical_2008} and {\em without} any forward or inverse model evaluations by exploiting the fact that samples from $p_b$ can be readily drawn using ancestral sampling i.e. by drawing samples of $\bs{u}$ from $p_u$ and propagating those with $\bs{b}_{NN}$. We note that finding this representation can become more diffucult in case $M$ is large but the complexity of the algorithms involved in general scales linearly with $M$ \citep{bishop2006pattern}.

By combining the (approximate prior) $p_{b,G}(\bs{b})$ above with the Gaussian likelihood $p(\bs{\hat{s}} \mid \bs{b})$ of \refeq{eq:likeb} we obtain an expression for the posterior $\tilde{p} (\bs{b} \mid \bs{\hat{s}} )$ using Bayes' theorem:
\be
\tilde{p} (\bs{b} \mid \bs{\hat{s}} ) \propto p(\bs{\hat{s}} \mid \bs{b})        p_{b,G}(\bs{b})
\ee
Due to the conjugacy of prior and likelihood, we can directly conclude that the (approximate) posterior is also a mixture of Gaussians \citep{bishop2006pattern}. Therefore, using expressions for the aforementioned likelihood/prior pair, we obtain a closed-form posterior $\tilde{p} (\bs{b} \mid \bs{\hat{s}} )$ on $\bs{b}$  of the form:
\be
 \tilde{p}(\bs{b}\mid\bs{\hat{s}})= \sum_{m=1}^M \tilde{w}_j ~\mathcal{N}(\bs{b} \mid \bs{\mu}_{b,m}, \bs{C}_{b,m})
 \label{eq:mixposterior}
  \ee
  where the mean $\bs{\mu}_{b,m}$ and covariance $\bs{C}_{b,m}$ of each mixture component can be computed as:
  \be
\begin{array}{l}
\bs{C}_{b,m}^{-1}=\sigma^{-2} \bs{Y}^T\bs{Y}+\bs{S}_{b,m}^{-1} \\
\bs{C}_{b,m}^{-1}\bs{\mu}_{b,m}=\sigma^{-2}\bs{Y}^T\bs{\hat{s}}+\bs{S}_b^{-1} \bs{m}_{b,m}
\end{array}
\ee
 The weights $\tilde{w}_m$ ($\sum_{m=1}^M \tilde{w}_m =1$) would be proportional to:
\be
\begin{array}{ll}
\tilde{w}_m  \propto w_m \mid \bs{D}_m \mid ^{-1/2} &\exp (-\frac{1}{2} (\hat{\bs{s}}-\bs{Y} \bs{m}_{b,m})^T \\ &\bs{D}_m^{-1} (\hat{\bs{s}}-\bs{Y} \bs{m}_{b,m}) )
\end{array}
\ee
where:
\be
  \bs{D}_m =\sigma^2 \bs{I} + \bs{Y} \bs{S}_{b,m} \bs{Y}^T
  \ee
Therefore inference tasks on the sought $bs{u}$  can be readily carried out by sampling $\bs{b}$  from the mixture-of-Gaussians posterior above  and propagating those samples through the inverse map $\bs{b}_{NN}^{-1}$ to obtain $\bs{u}$-samples. 
We note that by employing a mixture of Gaussians with sufficient components $M$,  one can approximate with arbitrary accuracy any non-Gaussian density as well as capture multimodal posteriors, a task that is extremely cumbersome with standard, Bayesian inference schemes  \citep{franck_multimodal_2017}.

\section{Numerical Illustrations}
\label{sec:num}
We applied the proposed framework to three examples, i.e. the  antiderivate operator , a reaction-diffusion PDE as well as a Darcy-type elliptic PDE.  In each of these cases,  we report the relative errors of forward and inverse maps (on test data) when trained  with  varying amounts of labeled and unlabeled training data. For the reaction-diffusion PDE and the Darcy-type elliptic PDE, we also use the proposed invertible-DeepONet-surrogate to solve pertinent Bayesian inverse problems. The code for the aforementioned numerical illustrations is available here\footnote{URL \url{https://github.com/pkmtum/Semi-supervised_Invertible_Neural_Operators}}. 
In Table \ref{tab:sum}, we summarize the most important dimensions for each of the following examples, namely $D$: the dimension of the PDE-input, $K$: the dimension of the observed PDE-output, $N_l$: number of labeled data (Equations (\ref{eq:lforw}), (\ref{eq:lback})), $N_u$: the number of unlabeled data (e.g. \refeq{eq:uinv}), $N_{res}$: the number of interior collocation points (\refeq{eq:lres}) and $N_{BC}$ the number of boundary collocation points (\refeq{eq:ubc}).

\begin{table*}
\begin{center}
\begin{minipage}{\textheight}
\begin{tabular}{|l|c|c|c|c|c|}
 \hline  
 & section \ref{sec:antiderivative} & section \ref{sec:rd} &  section \ref{sec:Darcy} & section \ref{sec:brd} & section \ref{sec:bdarcy1}  \\
 \hline
 $D$ & $100$ & $100$  & $64$  & $100$    &  $64$\\
 $K$ & $200$ & $200$  & $3844$ & $25,100$  & $1922,3844$ \\
 $N_l$ & $10^2,10^3,10^4$ &   $0, 500, 5000$ & $10^3$ & $500$ & $5000$ \\
 $N_u$  & $10^4$ & $5000$ & $10^3$ & $10^4$ &  $5000$ \\
 $N_{res}$  & $200$ & $200$ & $3844$ & $200$ & $200$ \\
 $N_{BC}$  & - &  $300$ & - & -  &  - \\
 \hline
\end{tabular}
\end{minipage}
\end{center}
\caption{Main dimensions for each numerical illustration}
 \label{tab:sum}
\end{table*}

\subsection{Anti-derivative Operator}
\label{sec:antiderivative}
As a first test case we considered the antiderivative operator on the interval $\xi  \in [0,1]$ with:\\
\begin{equation}
    \frac{d s( \xi)}{d \xi} = u(\xi) \text{  with   } s(0)=0
\end{equation}
i.e. when the input $u$ corresponds to the right-hand-side of this ODE and the operator $\mathcal{G}(u)$ that we attempt to approximate is simply the integral operator $\mathcal{G}(u)(\xi)=\int_0^{\xi} ~u(t)~dt$.
We generated $N_u=10000$   unlabeled training data by sampling  inputs $u$  from a Gaussian process with zero mean and exponential quadratic covariance kernel with a length scale $\ell=0.2$. Their values at  the same $D=100$  uniformly-distributed   locations in $[0,1]$ were recorded. We subsequently randomly choose $N_{res}=200$ collocation points to evaluate the residuals (see \refeq{eq:lres}).\\

Moreover, we used up to $N_l=10000$ labeled training data, for which the inputs were generated as for the unlabeled training data, and the outputs were obtained by solving the ODE above and  evaluating it at $K=200$ randomly chosen points.
We trained the invertible DeepONet  on $N_u=10000$ unlabeled training data with a batch size of $100$. In each batch we  added $1$, $10$ or $100$ labeled training data points per batch (i.e. $N_l=100, 1000, 10000$ respectively in Equations (\ref{eq:lforw}), (\ref{eq:lback})). A minimum of  one labeled datapoint is required in order to set the initial  condition correctly as  we did not enforce  this separately in the unlabeled loss part. With regards to the architecture of the networks used, we employed a MLP with four layers and 100 neurons each for the trunk network and 6 RealNVP building blocks for the branch network which were parametrized by a two-layered MLP. Variations around these values in the number of neurons, layers were also explored (in the subsequent examples as well) and did not impact significantly the performance.

Using the ADAM optimizer and an initial learning rate of $10^{-3}$, we run the model training for $4\times 10^4$  iterations with an exponential learning rate decay with rate $0.9$ every $1000$ iterations. As test data, we used $1000$ new (i.e. not included in the training data) input-output pairs  and compared the predicted forward and inverse solutions with the actual ones. The results obtained in terms of the relative errors are summarized in  Table \ref{tab:antiderivative}.
\begin{table*}
\sidewaystablefn%
\begin{center}
\begin{minipage}{\textheight}
    \begin{tabular}{|c|c|c|c|}\hline
         labeled data [\%] & 1 &10 &100  \\\hline
         relative error  $s$ (forward map)& $0.0152 \pm 0.0151$ & $0.00791 \pm 0.00799$ & $0.00728 \pm 0.00797$ \\\hline
         relative error  $u$ (inverse map) & $0.0371 \pm 0.0241$ & $0.034 \pm 0.024$ & $0.0215 \pm 0.0153$ \\\hline
    \end{tabular}
    \caption{Relative test errors and their standard deviations depending on the amount of labeled training data for the\\ anti-derivative operator. The percentage of labeled data is the amount of data used in comparison to unlabeled training\\ data, e.g. in the $10\%$ case we used ten times more unlabeled training data whereas in the $100\%$ case the\\ amount of labeled and unlabeled training data was the same.}
    \label{tab:antiderivative}
\end{minipage}
\end{center}
\end{table*}

\begin{figure*}[h]
    \centering
    \includegraphics[trim=0 0 430 0, clip,scale=0.275]{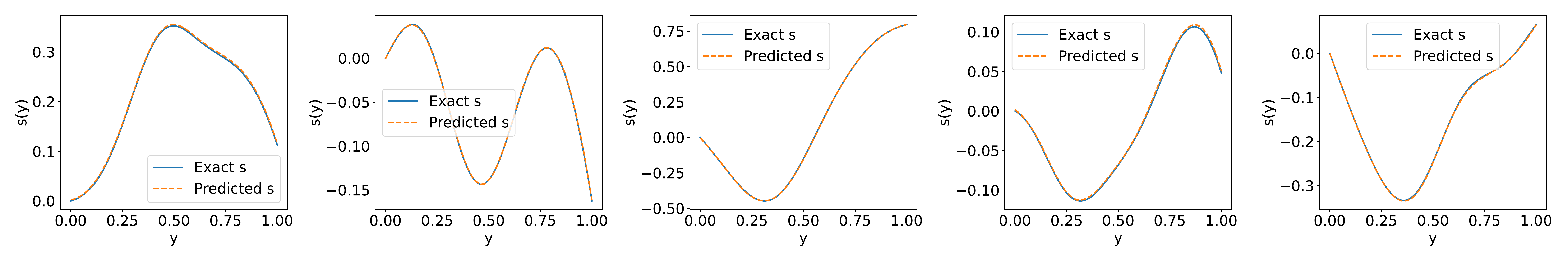}
    \caption{Forward map - Comparison of the true PDE-output/solution $s$ (given a PDE-input $u$) with the one predicted by the proposed invertible DeepONet and for the anti-derivative operator}
    \label{fig:adf}
\end{figure*}

\begin{figure*}[h]
    \centering
    \includegraphics[trim=430 0 0  0, clip,scale=0.275]{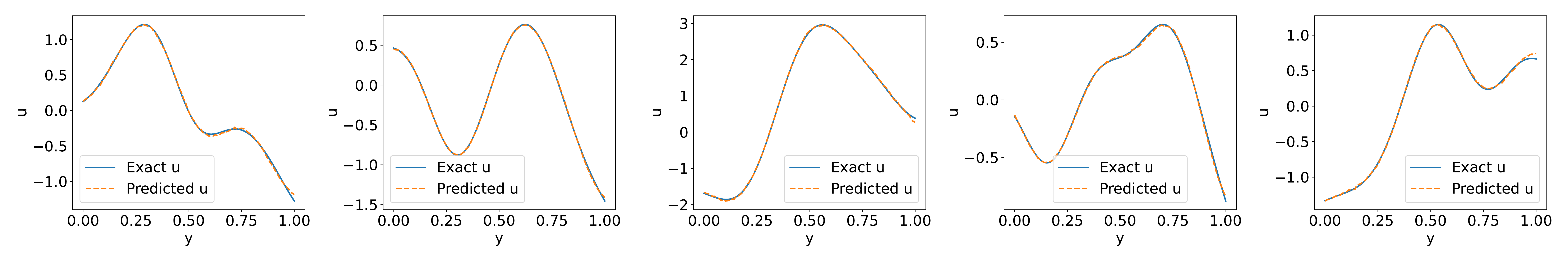}
    \caption{Inverse map - Comparison of the true PDE-input  $u$ (given the PDE-output/solution  $s$) with the one predicted by the proposed invertible DeepONet and for the anti-derivative operator}
    \label{fig:adb}
\end{figure*}

The  error values  indicate that both the forward as well as the inverse maps  are very well approximated by the proposed  invertible DeepONet. The addition of more labeled training data results in even lower errors especially for the inverse map for which  the relative   error is decreased from almost $\sim 4\%$ to $\sim 2\%$.\\
In order to visualize the results we plot for four randomly-chosen test cases  the predictions (when trained with $10\%$ labeled data) of both the forward (Figure \ref{fig:adf}) and inverse (Figure \ref{fig:adb}) operator.  In all cases, the predictions are indistinguishable from the reference functions.


In Appendix \ref{sec:appendixA} we include additional results  for this problem  with varying amounts of unlabeled and labeled training data in order to further show their influence.

\subsection{Reaction-Diffusion dynamics}
\label{sec:rd}
The second illustrative example involves the reaction-diffusion equation on the space-time domain $ \bs{\xi}=(x,t) \in [0,1]\times[0,1]$:
\begin{equation}
    \frac{ \partial s}{\partial t} = D_s \frac{ \partial^2 s}{\partial x^2} + k s^2 + u(x)
\end{equation}
Here, $D_s=0.01$ is the diffusion constant,  $k=0.01$  the reaction rate and the source-term  $u(x)$  is chosen to be the PDE-input. We used zero values  as initial conditions and  boundary conditions. We generated random source terms by sampling from a Gaussian process with zero mean and and exponential quadratic covariance kernel with a length scale $\ell=0.2$ which were then evaluated at $D=100$ uniformly  distributed points over $[0,1]$. 
The PDE was subsequently solved using an implicit Finite-Difference scheme and evaluated at $200$ randomly chosen points to generate the labeled training data.

We trained our model with  $N_u=5000$ unlabeled  data  which were processed in batches of $100$ samples and to which  varying  amounts of labeled  data were added. Since  for this problem the boundary conditions were enforced separately, the amount of labeled training data used could also be zero. All unlabeled training data points were evaluated at $N_{res}=200$ randomly selected collocation points.
With regards to the network architecture, we employed a MLP with five layers and 100 neurons each for the trunk network and 3 RealNVP building blocks for the branch network which were parametrized by a three-layered MLP. Using the ADAM optimizer and an initial learning rate of $10^{-3}$, we run the model training for $12\times 10^4$ iterations with an exponential learning rate decay with rate $0.9$ every $2000$ iterations. For our test dataset, we generated $1000$ new (unseen) source terms ${u}$ and corresponding solutions ${s}$. A summary of the relative errors obtained is contained  in Table \ref{tab:rd}.
\begin{table*}
\sidewaystablefn%
\begin{center}
\begin{minipage}{\textheight}
    \begin{tabular}{|c|c|c|c|}\hline
         labeled data [\%] & 0 & 10 & 100 \\\hline
         relative error for s & $0.00925 \pm 0.00492$ & $0.0105 \pm 0.00519$ & $0.00813 \pm 0.00445$  \\\hline
         relative error for u & $0.024 \pm 0.01021$ & $0.0184 \pm 0.00578$ & $0.0162 \pm 0.00592$  \\\hline
    \end{tabular}
    \caption{Relative errors on test data  depending on the amount of labeled training data for the reaction-diffusion case.}
    \label{tab:rd}
\end{minipage}
\end{center}
\end{table*}

\begin{figure*}[h!]
    \includegraphics[scale=0.25]{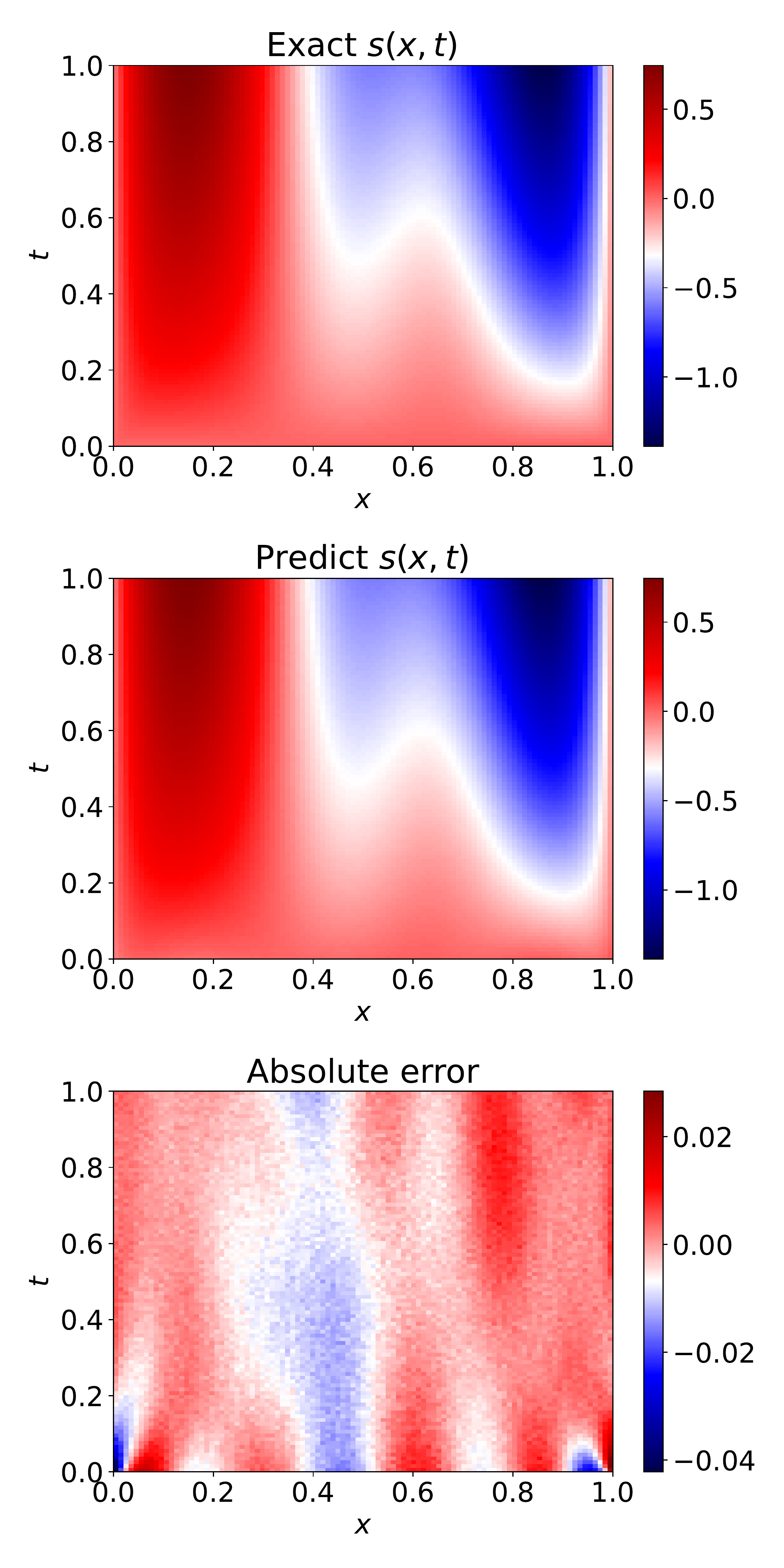}
    \includegraphics[scale=0.25]{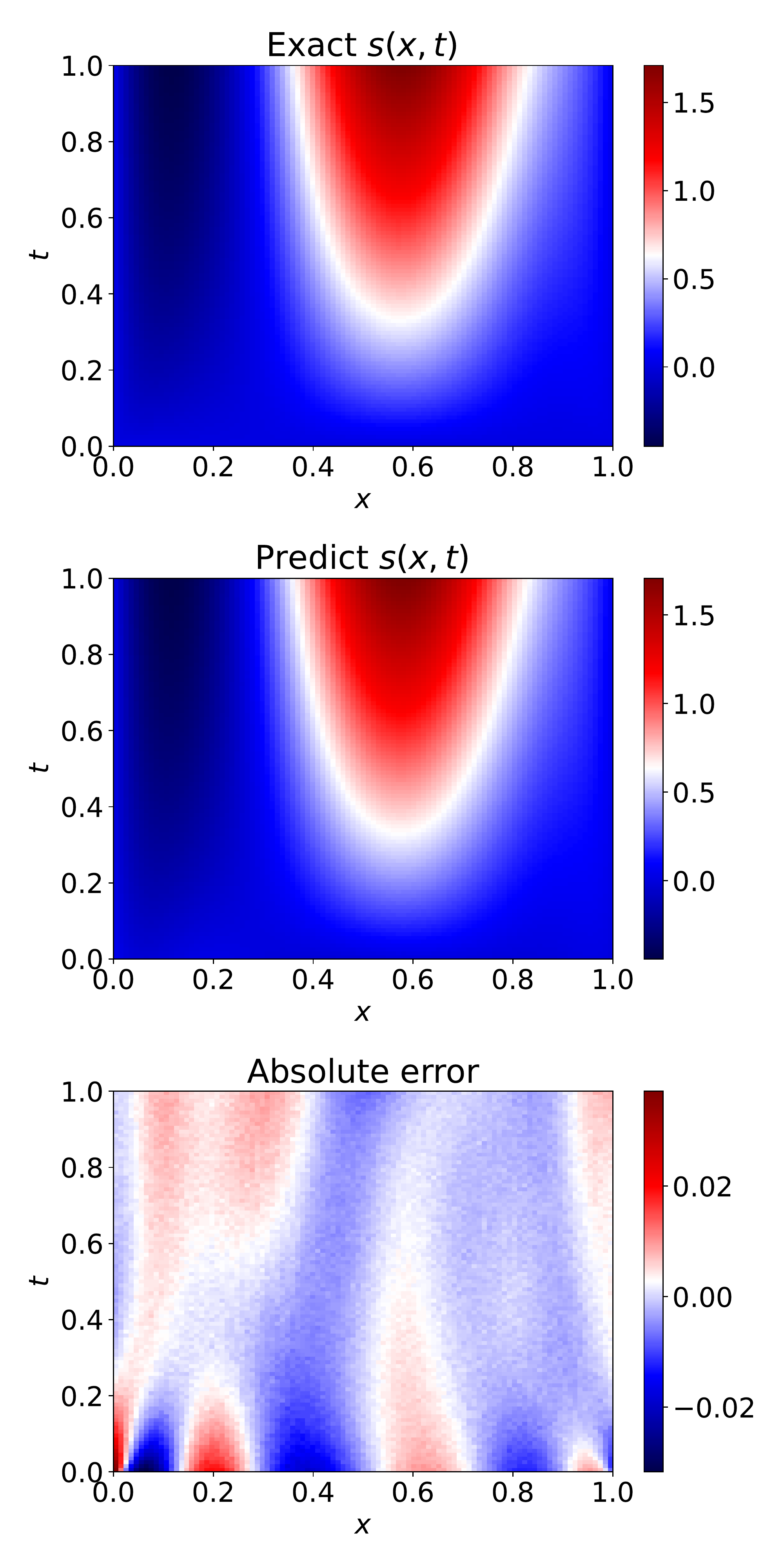}
    \includegraphics[scale=0.25]{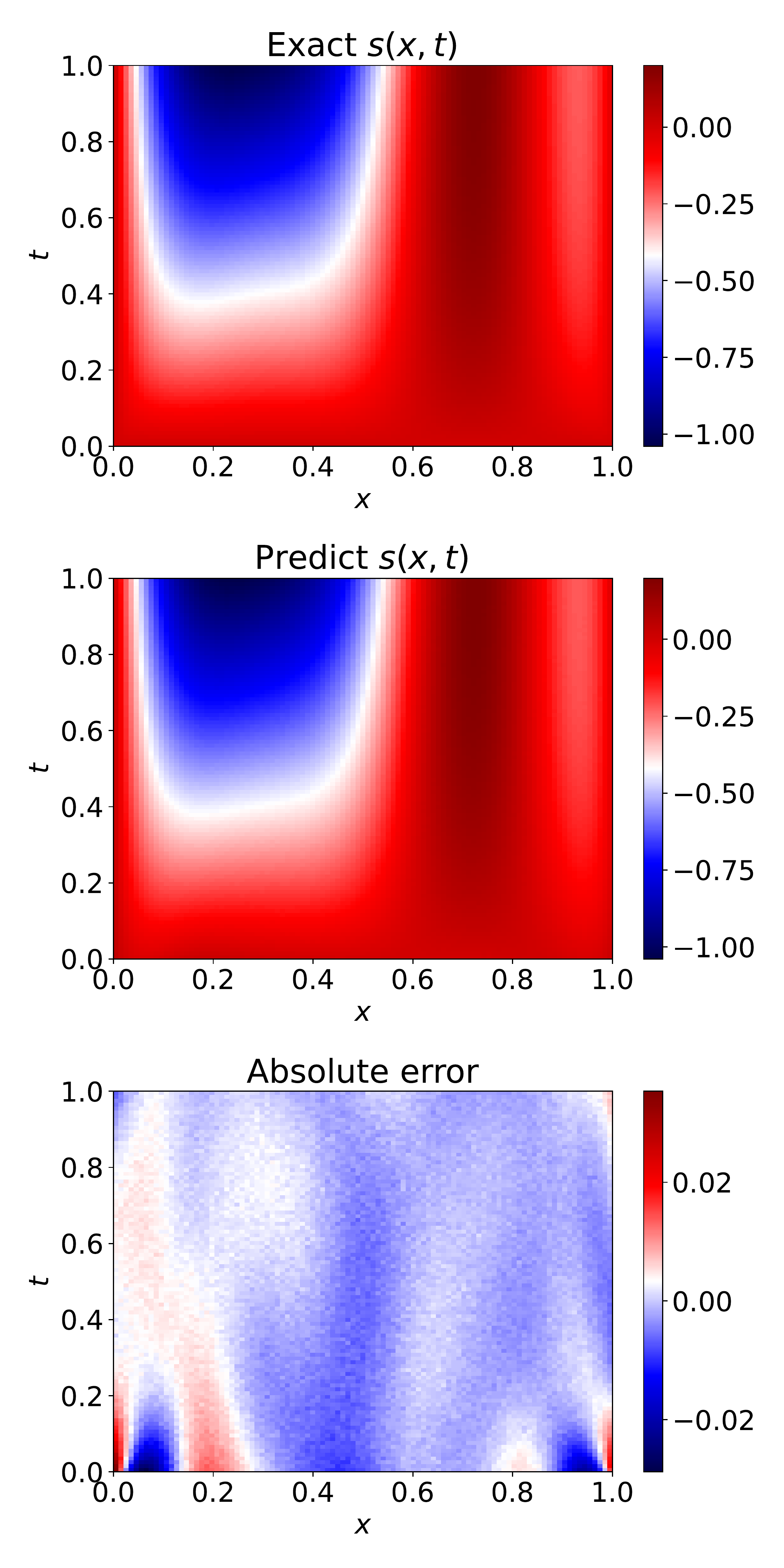}
    \caption{Forward map - Comparison of the true PDE-output/solution $s$ (given the PDE-input $u$) with the one predicted by the proposed invertible DeepONet and for Reaction-Diffusion PDE.}
    \label{fig:RD_s}
\end{figure*}

\begin{figure}[h!]
    \centering
    \includegraphics[scale=0.3]{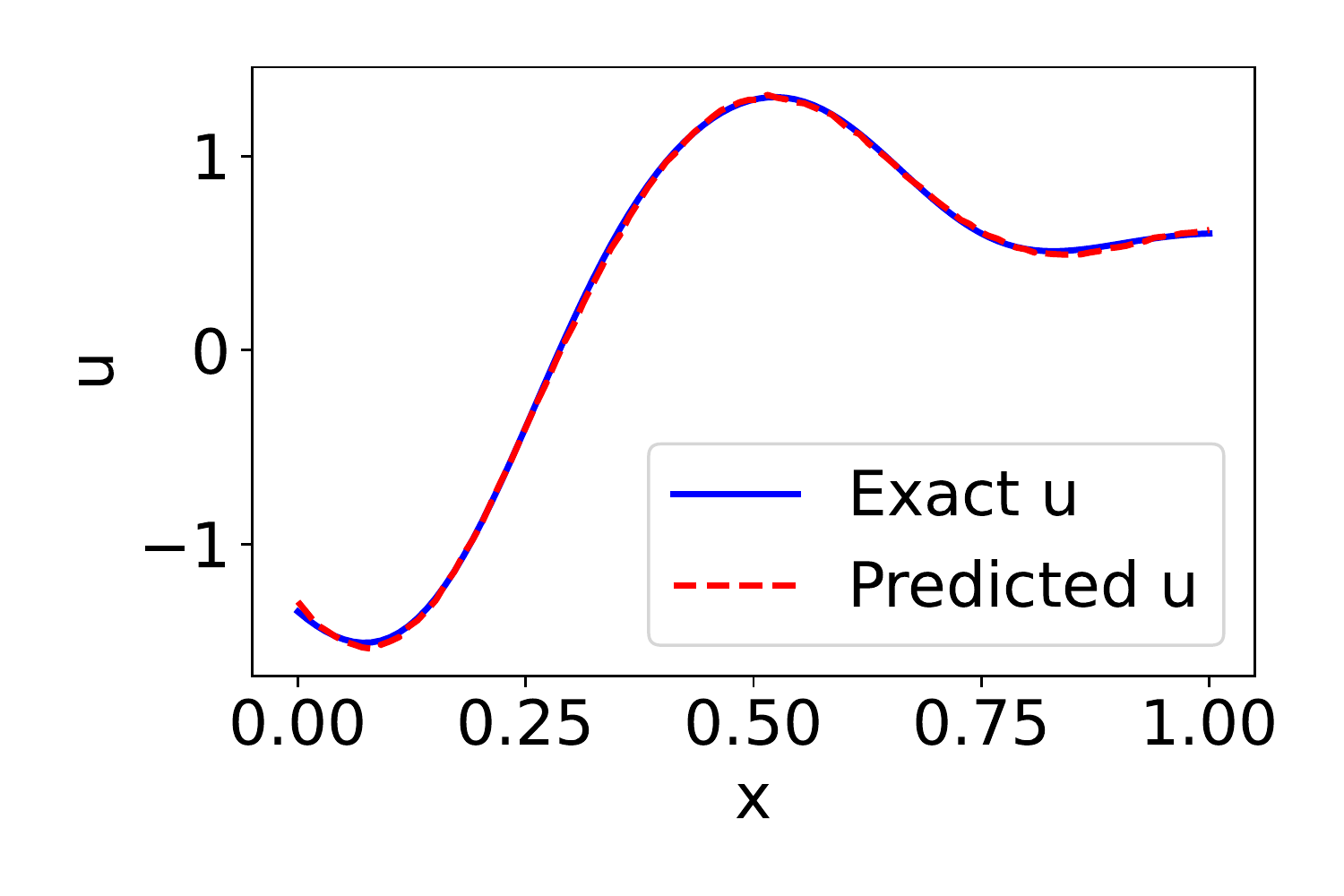}
    \includegraphics[scale=0.3]{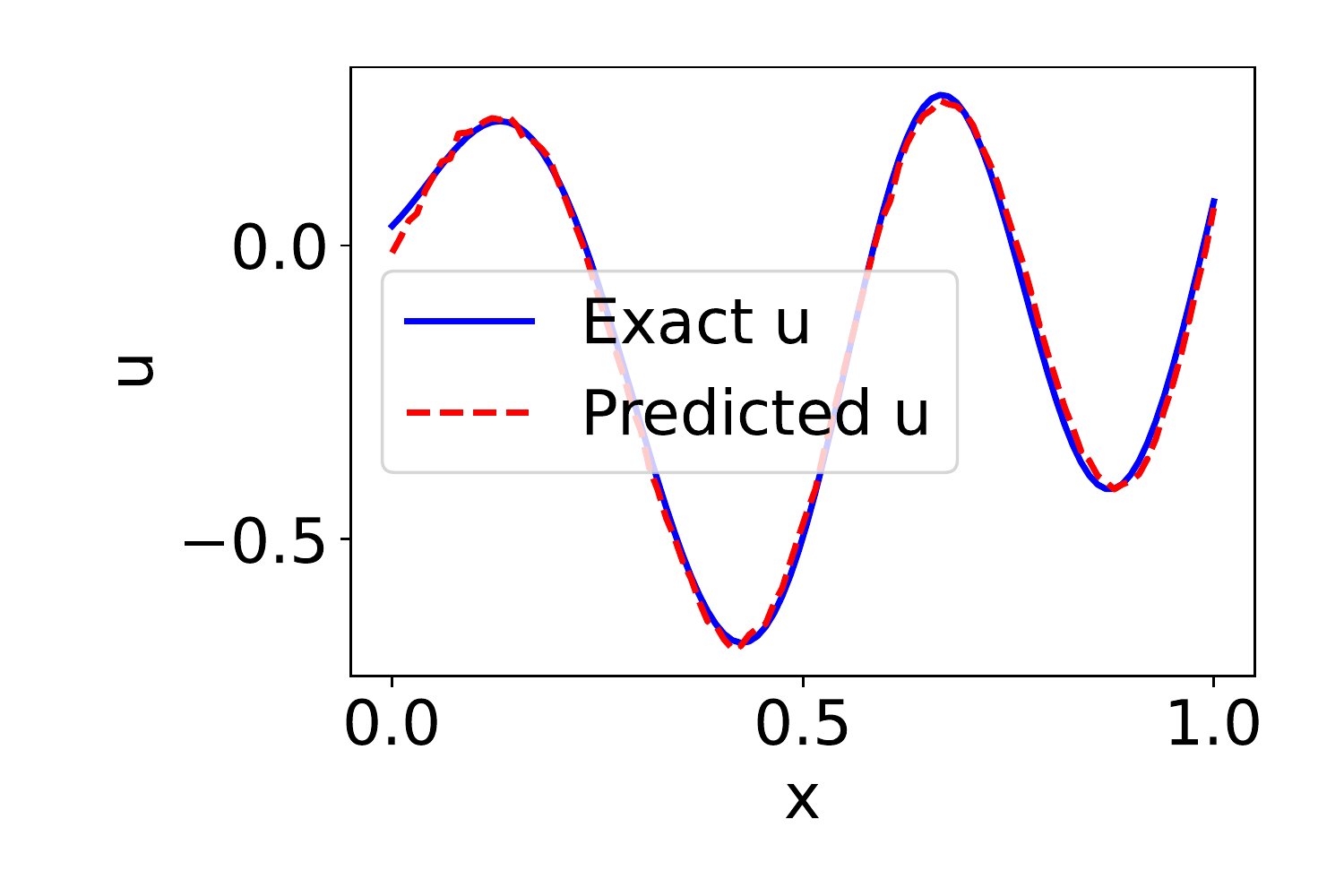}
    \includegraphics[scale=0.3]{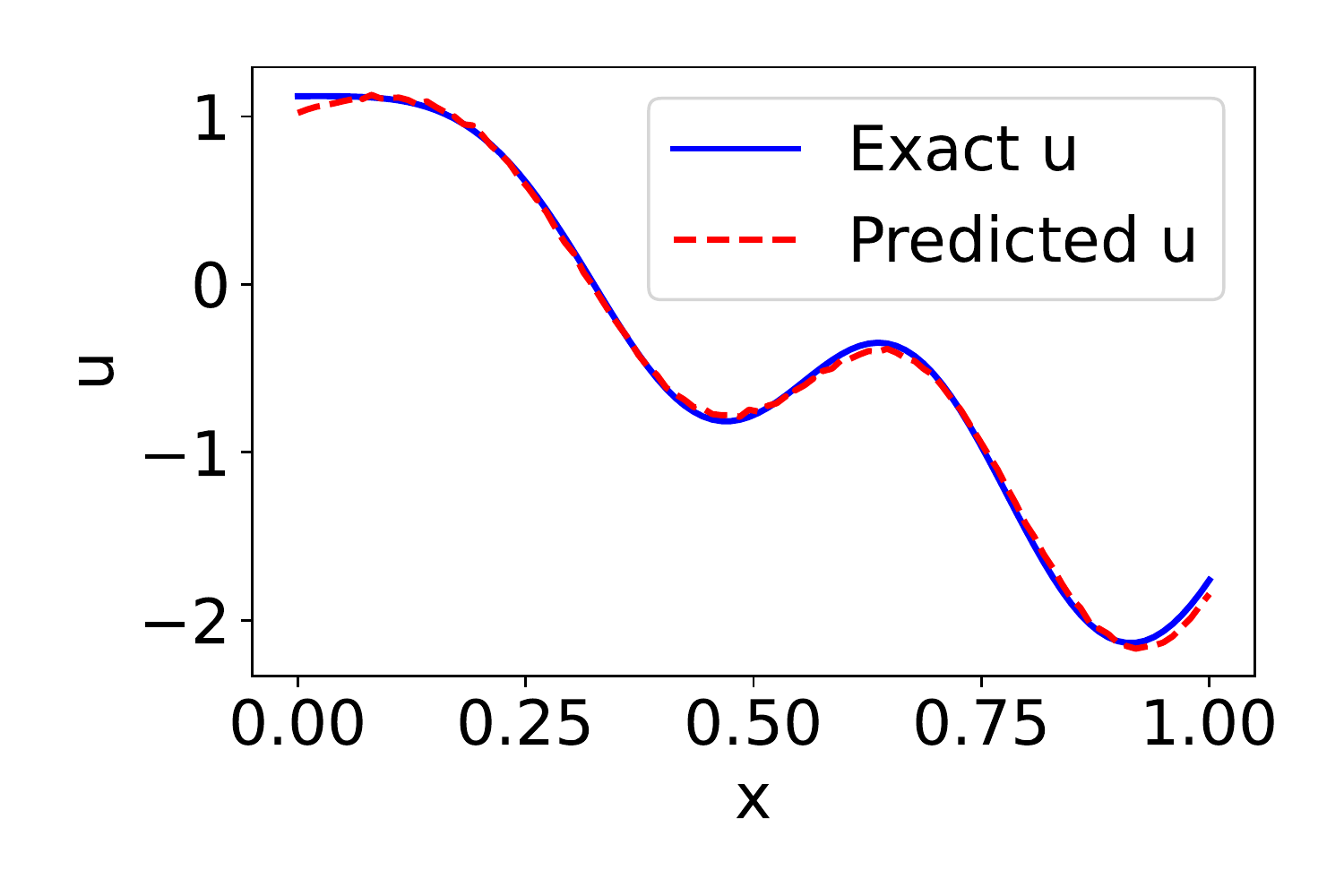}
    \caption{Inverse map - Comparison of the true PDE-input  $u$ (given the PDE-output/solution  $s$) with the one predicted by the proposed invertible DeepONet and for Reaction-Diffusion PDE.}
    \label{fig:RD_u}
\end{figure}

We note that again for all three settings we achieve very low error rates, which decrease as the amount of labeled training data increases. In Figure \ref{fig:RD_s} and \ref{fig:RD_u} we show the predictions (trained with $500$ i.e. $10\%$  labeled data)  of both forward and inverse map for three  randomly chosen test cases.


\subsection{Flow through porous media}
\label{sec:Darcy}
In the  final example we considered the Darcfy-flow elliptic PDE in the two-dimensional domain  $\bs{\xi}=(x_1,x_2) \in [0,1]^2$
\begin{equation}
    \nabla \cdot (u( \bs{\xi}) \nabla s(\bs{\xi}))= 10
    \label{eq:darcy}
\end{equation}
where the PDE-input $u$ corresponds to the permeability field.
We assumed  zero values for the solution  $s$ along all boundaries 
 which we a-priori incorporated in our operator approximation by  multiplying the DeepONet expansion  in \refeq{eq:donet} with the polynomial  $x_1(1-x_1)~x_2(1-x_2)$. 
We used $N_u=1000$ unlabeled training data points with $N_{res}=3844$ collocation points (\refeq{eq:lres}) during training and added either no labeled training data at all (i.e. $N_l=0$) or $N_l=1000$.
In order to obtain the latter we solved \refeq{eq:darcy} with  the Finite Element  library  FEniCS \citep{logg2012automated} on a $128\times 128$ mesh with linear elements and evaluated the solution at $3844$ regularly distributed points.
We represent the  PDE-input  $u$ as follows\footnote{We employ this expansion for the logarithm of $u$ in order to ensure that the resulting  permeability field is positive}:
\begin{align}
\ln(u)=&\sum_{f_1=1}^4 \sum_{f_2=1}^4   c_{f_1,f_2,1} \sin(f_1 x_1) \cos(f_2 x_2) \nonumber\\
+&c_{f_1,f_2,2}\sin(f_1 x_1) \sin(f_2 x_2) \nonumber\\
+&c_{f_1,f_2,3}\cos(f_1 x_1) \sin(f_2 x_2) \nonumber\\
+&c_{f_1,f_2,4}\cos(f_1 x_1) \cos(f_2 x_2)
\label{eq:features}
\end{align}
using  $64$ feature functions and corresponding coefficients $c$. 
In order to generate the training data, we sampled each of the aforementioned  $64$ coefficients from a uniform  distribution in $[0,1]$. 
In this example the $64$-dimensional vector of the $c$'s serves as the input in the branch network (i.e. $D=64$). With the help of the $c$'s and of \refeq{eq:features}, one can reconstruct the full permeability field.

With regards to the  network architecture, we employed a MLP with five layers and 64 Neurons each for the trunk network and 3 RealNVP building blocks for the branch network which were parametrized by a three-layered MLP. Using the ADAM optimizer and an initial learning rate of $10^{-3}$, we run the model training for $10^5$ iterations with an exponential learning rate decay with rate $0.9$ every $2000$ iterations.
We tested the trained model on $2500$ unseen test data and obtained the results in Table \ref{tab:Darcy_features}. As in the previous examples, the inclusion of labeled data significantly improves the predictive accuracy of the trained model. For the case without data the predictive accuracy of the forward map is slightly lower but the accuracy in  the inverse map is comparably low. The addition of labeled data  improves the predictive accuracy for both  maps.

\begin{table}[h]
    \centering
    \begin{tabular}{|c|c|c|c|}\hline
         labeled data [\%] & 0 & 100\\\hline
         relative error for s & $0.0134 \pm 0.00509$ & $0.0245 \pm 0.0108 $  \\\hline
         relative error for u & $0.235 \pm 0.137 $  & $0.0566 \pm 0.0198$   \\\hline
    \end{tabular}
    \caption{Relative errors on test data depending on the amount of labeled training data for the Darcy example with feature coefficients as inputs.}
    \label{tab:Darcy_features}
\end{table}

\begin{figure*}
\centering
    \includegraphics[scale=0.38]{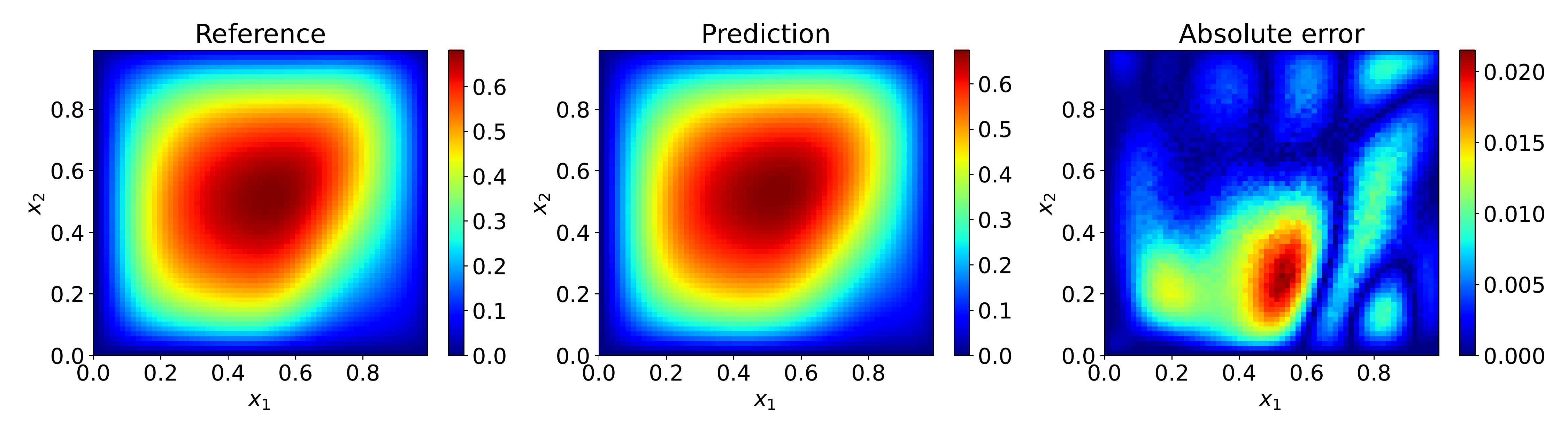}
    \includegraphics[scale=0.38]{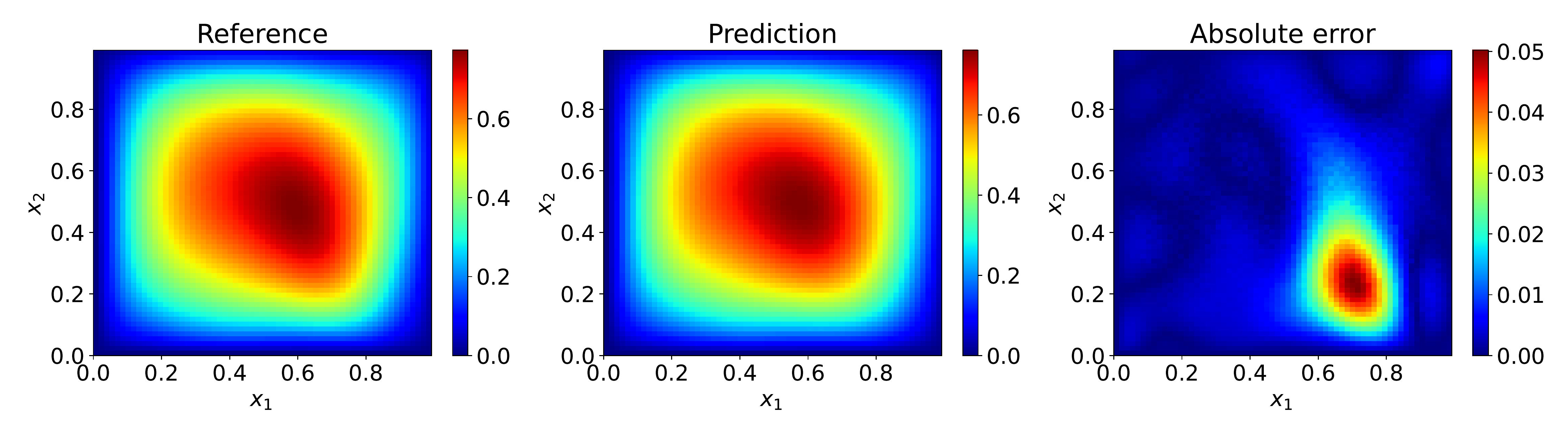}
    \caption{Forward map - Comparison of the true PDE-output/solution $s$ (given feature coefficients as the PDE-input $u$) with the one predicted by the proposed invertible DeepONet and for Darcy-type PDE.}
    \label{fig:Darcy_f_s}
\end{figure*}

\begin{figure*}
\centering
    \includegraphics[scale=0.38]{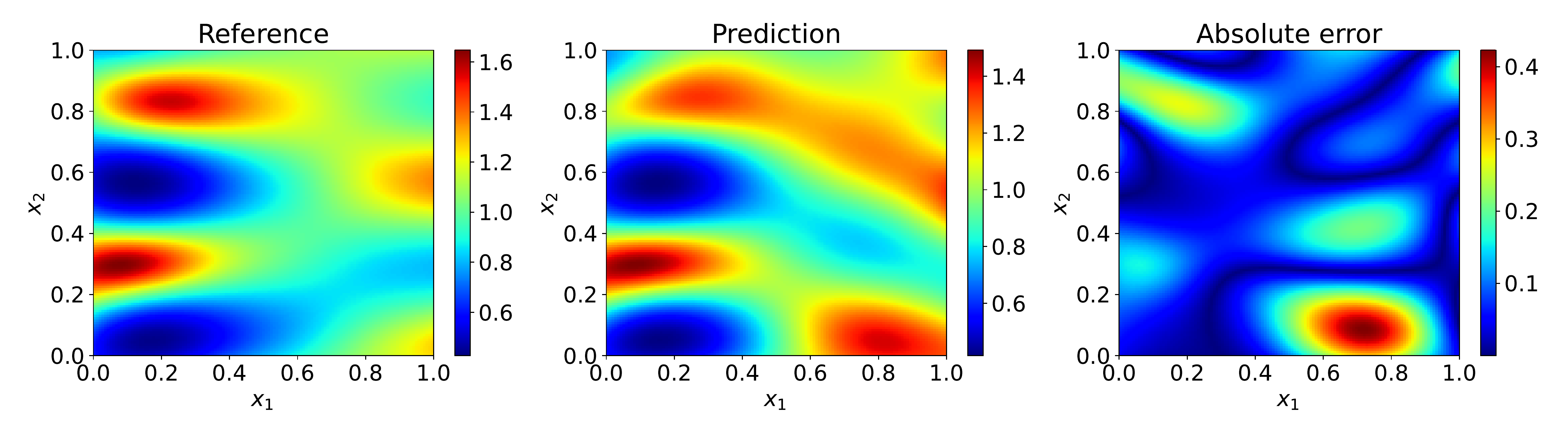}
    \includegraphics[scale=0.38]{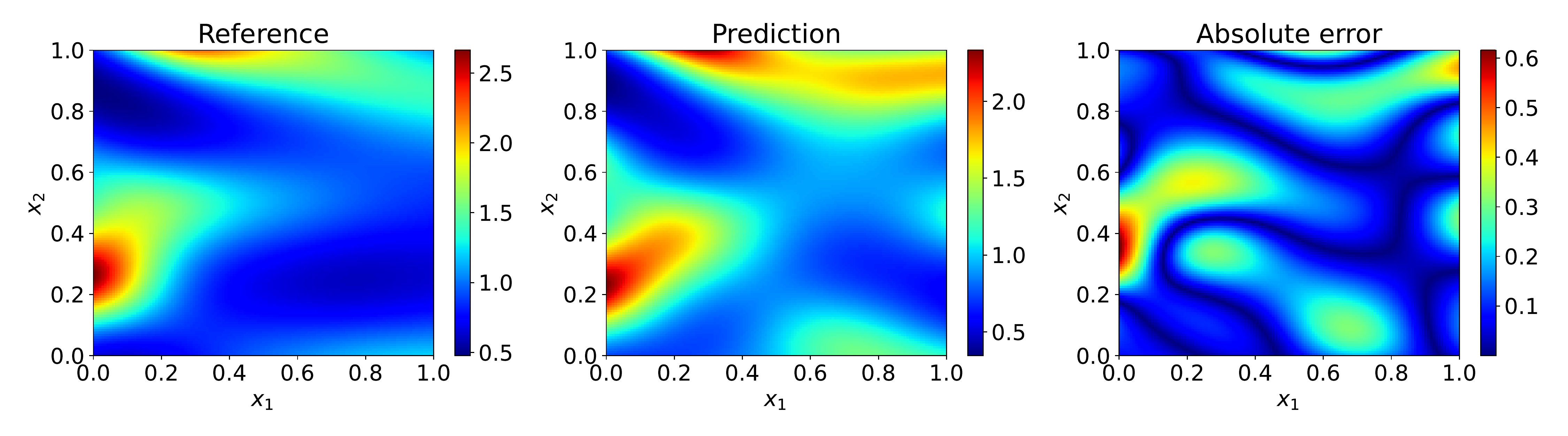}
    \caption{Inverse map - Comparison of the reconstructed PDE-input (given the PDE-output/solution  $s$) with the one predicted by the proposed invertible DeepONet and for Darcy-type PDE with zero labeled training data.}
    \label{fig:Darcy_f_woD_u}
\end{figure*}

\begin{figure*}
\centering
    \includegraphics[scale=0.38]{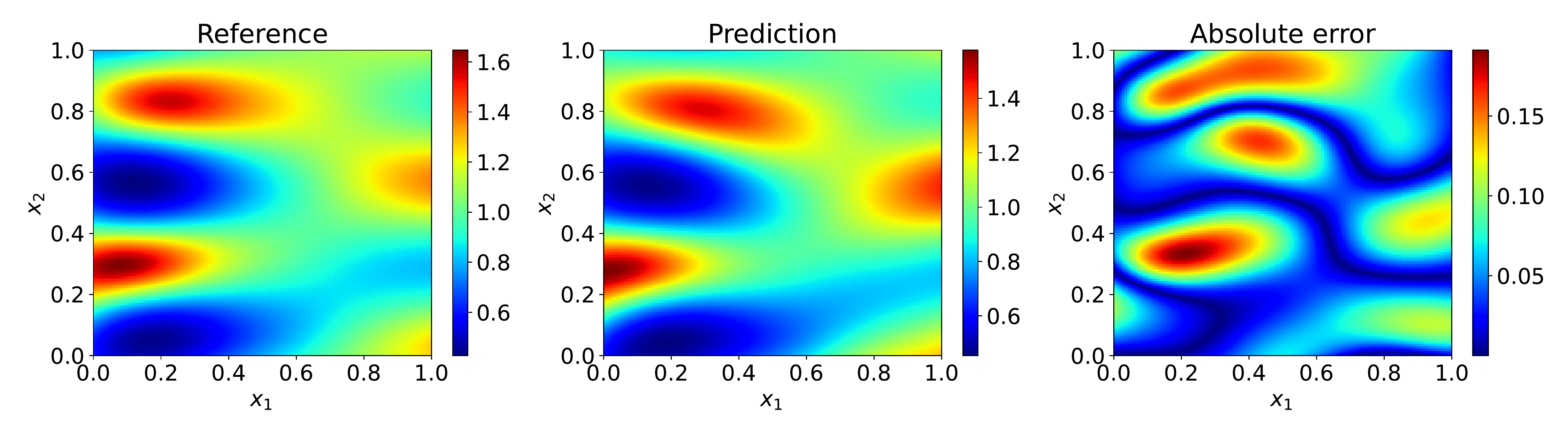}
    \includegraphics[scale=0.38]{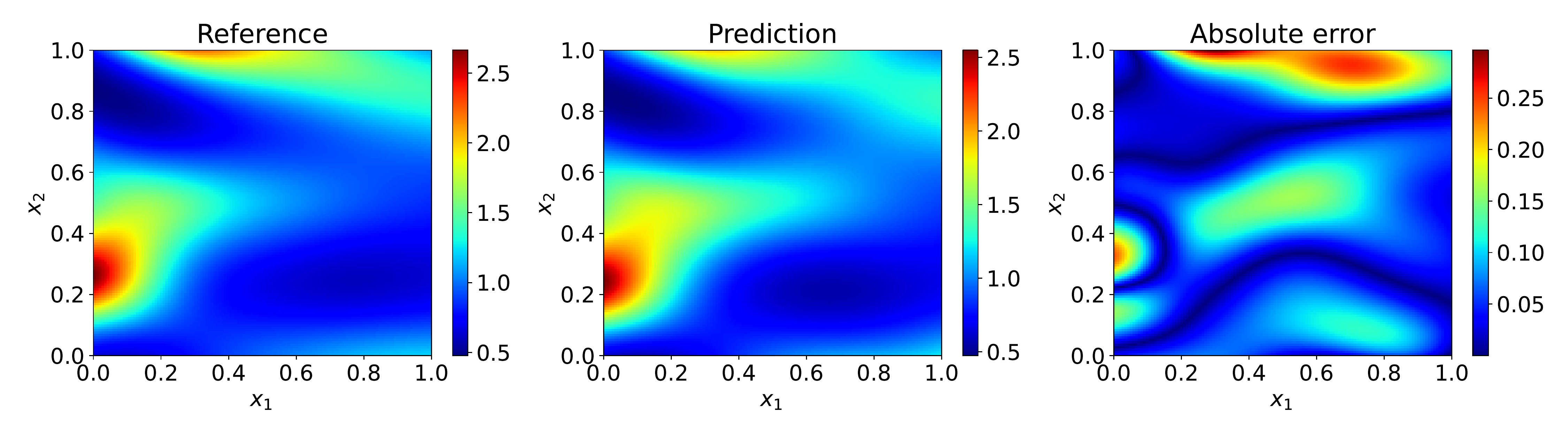}
    \caption{Inverse map - Comparison of the reconstructed PDE-input (given the PDE-output/solution  $s$) with the one predicted by the proposed invertible DeepONet and for Darcy-type PDE.}
    \label{fig:Darcy_f_u}
\end{figure*}

In Figure \ref{fig:Darcy_f_s} we compare the reference solution for two illustrative test cases with the the forward map learned  with labeled training data. As suggested by the cumulative results  in  Table \ref{tab:Darcy_features} the two predictions are very close to the reference and the accuracy is very high.
In Figures \ref{fig:Darcy_f_woD_u} (without labeled training)  and \ref{fig:Darcy_f_u}  (with labeled training) the results for two illustrative inverse test cases are shown.

While locally the error can be significant, the main characteristics of the PDE-input field $u$ can be captured.


We discuss in the next section the case where the input permeability field $u$ is not represented with respect to some feature functions but rather as a discretized continuous field. 

\subsubsection{Coarse-grained (CG) input parameters}
\label{sec:Darcy_CG}

In this sub-case, we modeled the permeability field $u$ with an exponentiated (to ensure positivity) Gaussian Process with mean zero and  exponential quadratic covariance with length scale $\ell=0.1$. The PDE was then again solved on a $128 \times 128$ FE mesh and the values of the solution $s$ were assumed to be observed at $3844$ regularly distributed points. We moreover sub-sampled the generated PDE input on a regular $8\times 8$ grid and its $D=64$ values represented the branch network input $\bs{u}$.
We  generated $N_u=1000$ unlabeled fields $u$ in total and used $N_{res}=3844$ collocation points (\refeq{eq:lres}) during training. We also trained the model with $N_l=1000$ labeled training data.

\begin{figure*}
\centering
    \includegraphics[scale=0.38]{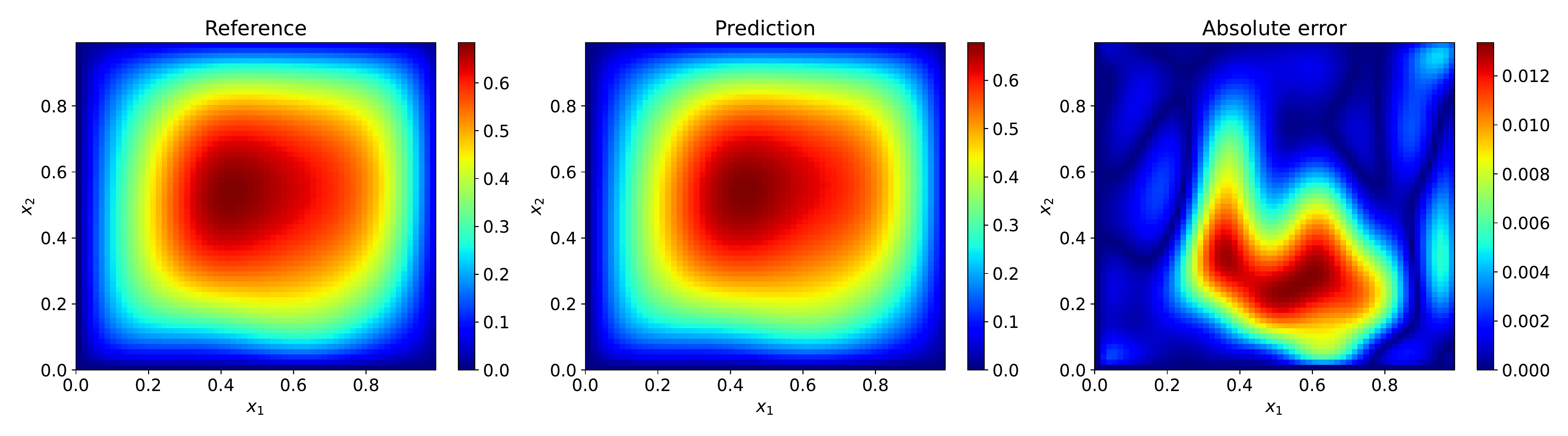}
    \includegraphics[scale=0.38]{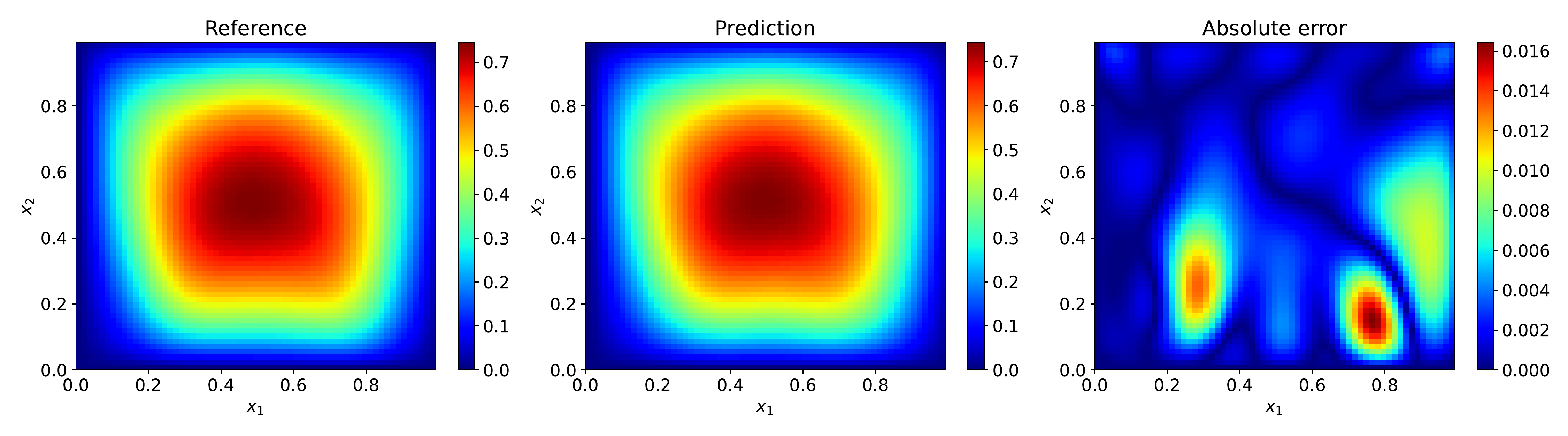}
    \caption{Forward map - Comparison of the true PDE-output/solution $s$ (given the coarse-grained PDE-input $u$) with the one predicted by the proposed invertible DeepONet and for Darcy-type PDE.}
    \label{fig:Darcy_CG_s}
\end{figure*}

The results obtained can be found in Table \ref{tab:darcy_cg}. The test data in  this table consists of $2500$ unseen, 
 discretized, permeability  fields and their respective solutions. The error rates are computed with respect to the coarse-grained reference input. As in the previous setting, we observe a significant improvement in the accuracy of the inverse map when labeled data are used in training. 
\begin{table}[h]
    \centering
    \begin{tabular}{|c|c|c|c|}\hline
         labeled data [\%] & 0 & 100\\\hline
         relative error for s & $0.0164 \pm 0.00712$ & $0.0164 \pm 0.00748$  \\\hline
         relative error for u & $0.121 \pm 0.041 $ & $0.0656 \pm 0.0168$\\\hline
    \end{tabular}
    \caption{Relative errors on test data depending on the amount of labeled training data for the Darcy example with coarse-grained input parameters}
    \label{tab:darcy_cg}
\end{table}
In Figure \ref{fig:Darcy_CG_s} we compare the reference solution for two illustrative test cases with the the forward map learned  with labeled training data. As suggested by the cumulative results in  Table \ref{tab:darcy_cg} the two predictions are very close to the reference and the accuracy is very high.

\begin{figure*}
\centering
    \includegraphics[scale=0.275]{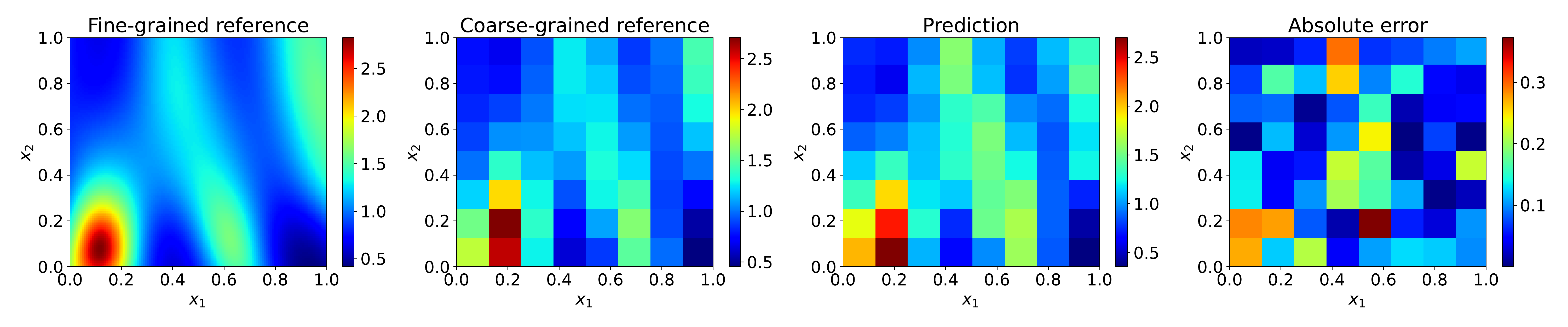}
    \includegraphics[scale=0.275]{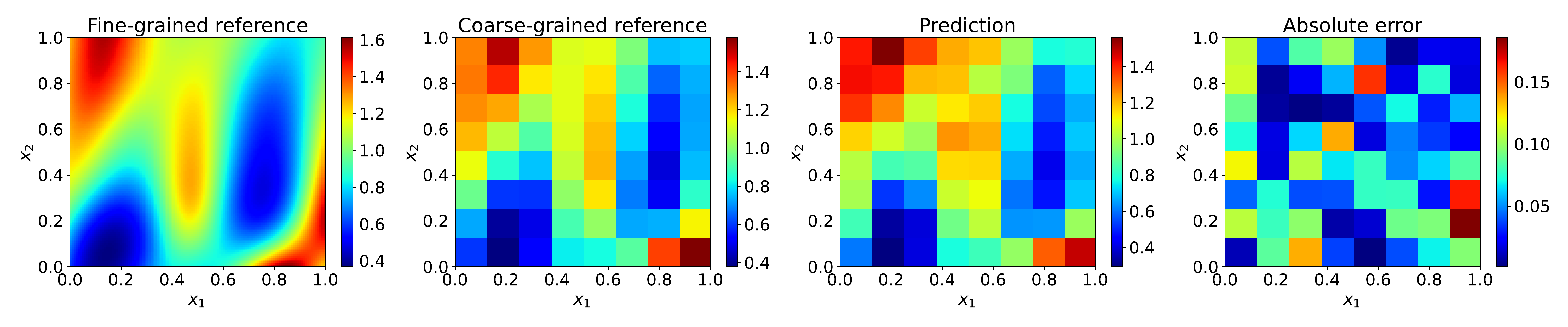}
    \caption{Inverse map - Comparison of the coarse-grained PDE-input $u$ (given the PDE-output/solution  $s$) with the one predicted by the proposed invertible DeepONet and for Darcy-type PDE with zero labeled training data.}
    \label{fig:Darcy_CG_woD_u}
\end{figure*}

\begin{figure*}
\centering
    \includegraphics[scale=0.275]{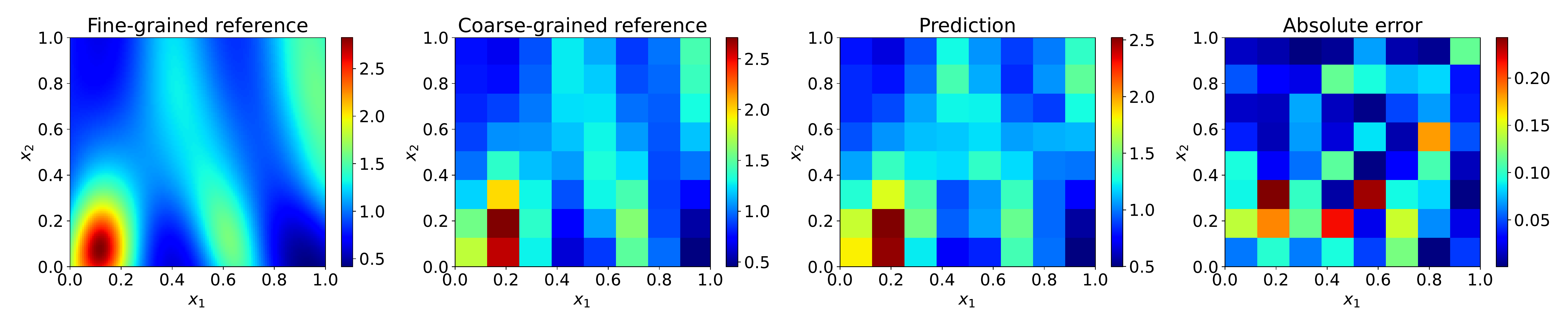}
    \includegraphics[scale=0.275]{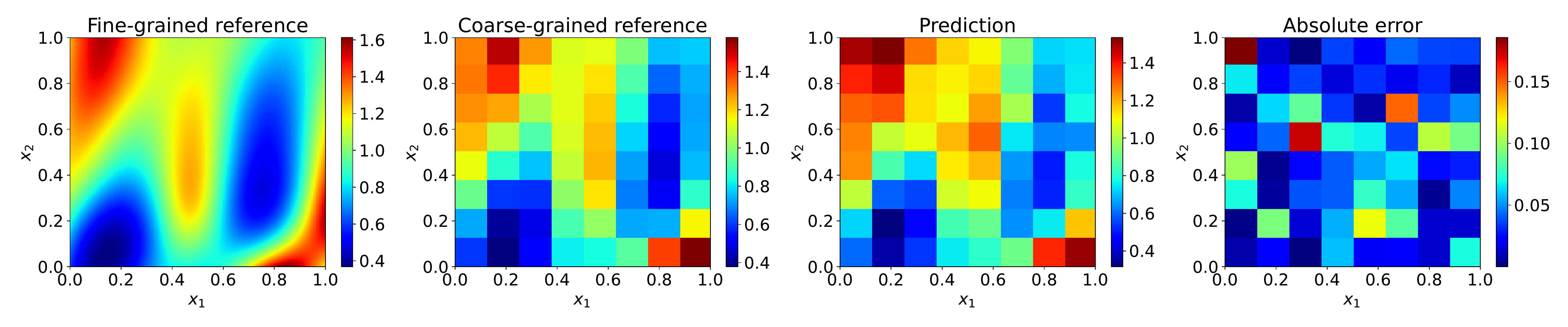}
    \caption{Inverse map - Comparison of the coarse-grained PDE-input (given the PDE-output/solution  $s$) with the one predicted by the proposed invertible DeepONet and for Darcy-type PDE.}
    \label{fig:Darcy_CG_u}
\end{figure*}

In Figures \ref{fig:Darcy_CG_woD_u} (without labeled training)  and \ref{fig:Darcy_CG_u}  (with labeled training) the results for two illustrative inverse test cases are shown.
We note again that the main features of the PDE-input's spatial variability are captured, despite the presence of localized errors.

\subsection{Bayesian Inverse Problems}
\label{sec:bayes}
In this section we demonstrate the utility of the invertible DeepONet proposed in the solution of   Bayesian inverse problems and in obtaining accurate approximations of the posterior without any need for additional reference model runs nor for any costly and asymptotically-exact  sampling. For each of the examples considered, only one observed output $\hat{\bs{s}}$ was assumed to be given. The variance of the observational noise $\sigma^2$ was assumed to be given although this could readily be inferred, especially if a conjugate inverse-Gamma prior was used for it. In this manner, any deviations from the actual posterior could be attributed to inaccuracies of the DeepONet-based surrogate. Errors due to the approximation of the prior with a mixture of Gaussians as in \refeq{eq:mixprior} can be made arbitrarily small by increasing the number of mixture components $M$.

\subsubsection{Reaction-Diffusion dynamics}
\label{sec:brd}
\begin{figure}[t!]
    \centering
    \includegraphics[scale=0.4]{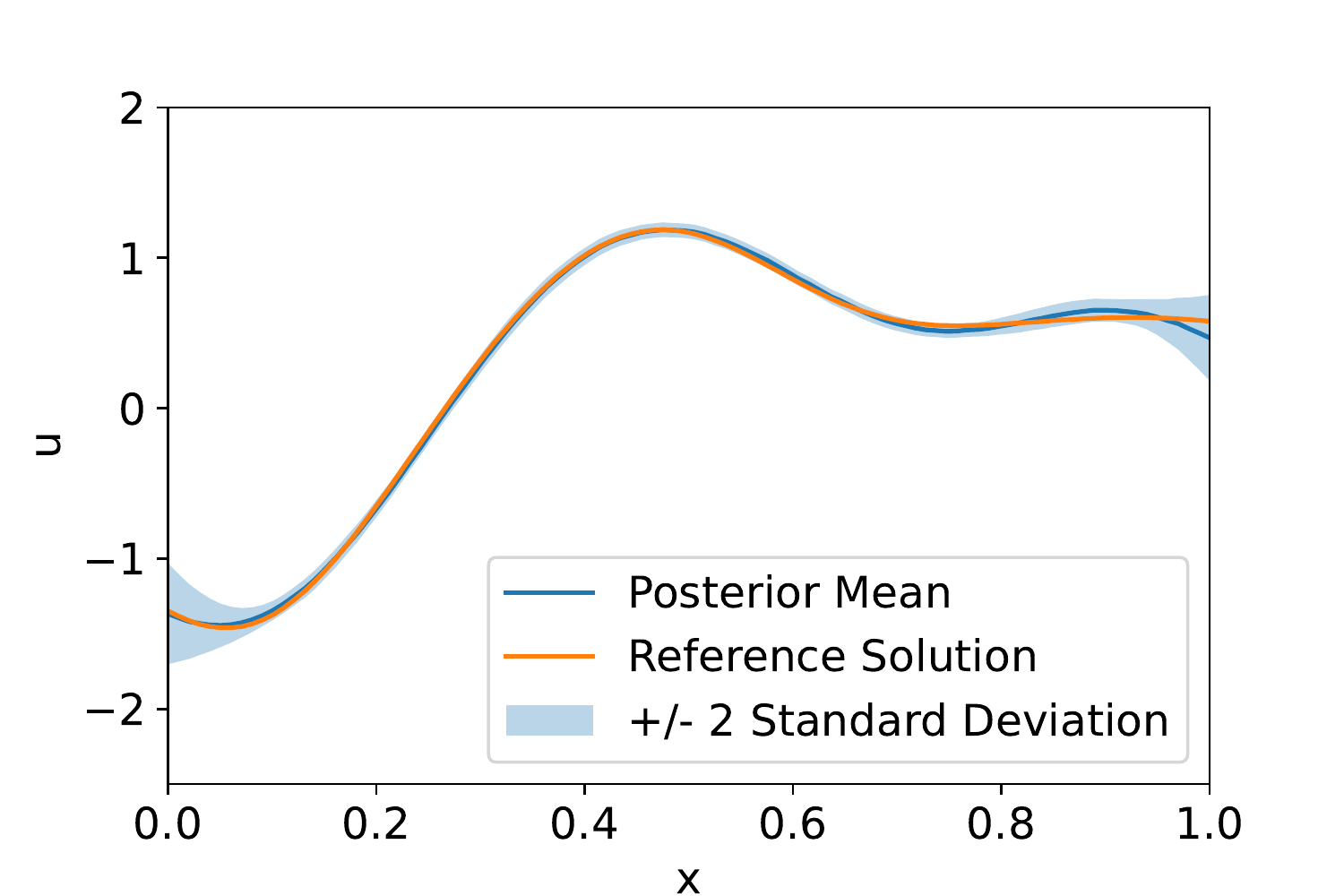}
    \includegraphics[scale=0.4]{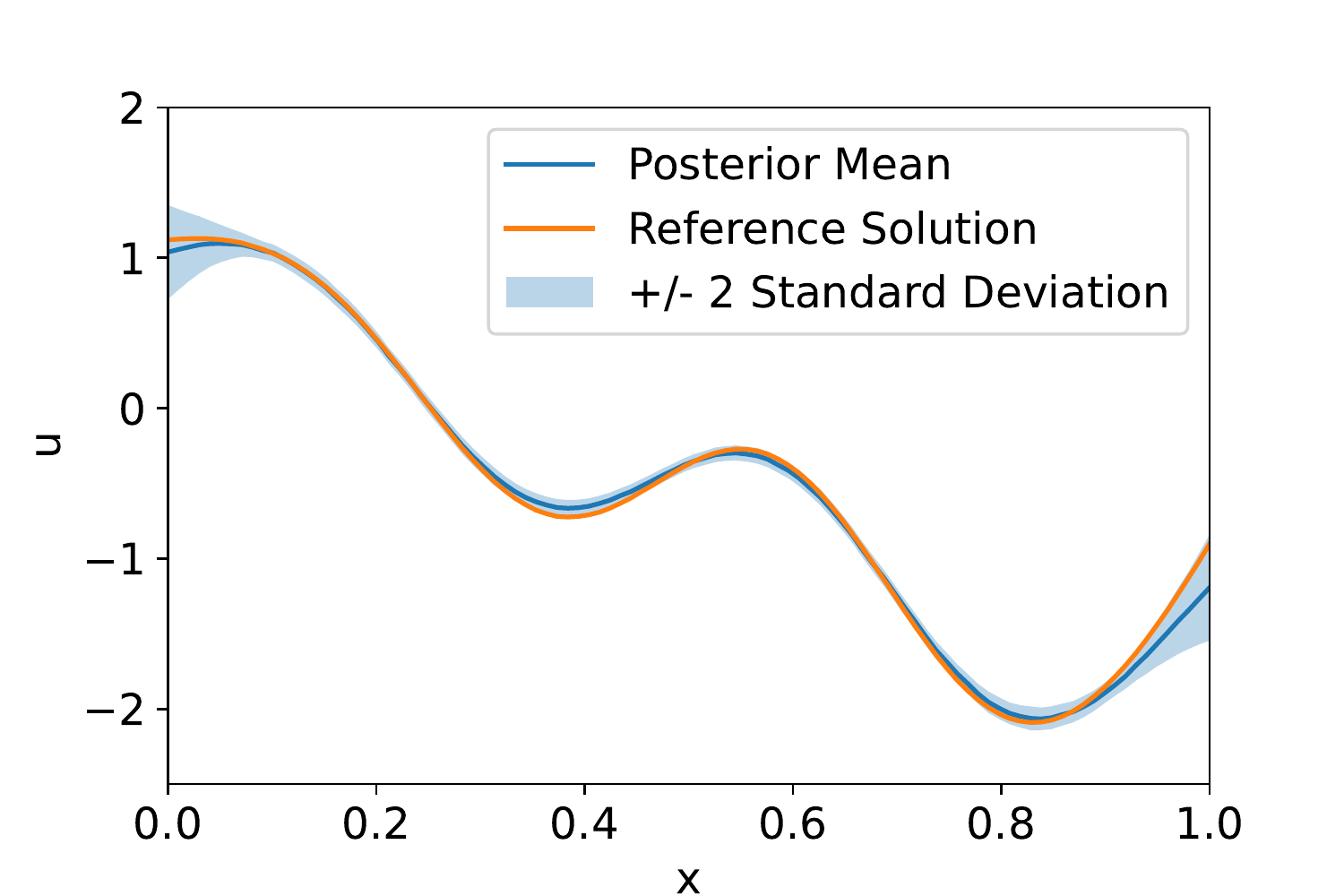}
    \caption{Bayesian Inverse Problem  for Reaction-Diffusion PDE: 100 observed data points with $\sigma^2=0.001$ and a 100-dimensional parametric input.}
    \label{fig:Inv_RD_1}
\end{figure}

\begin{figure}[h]
    \centering
    \includegraphics[scale=0.4]{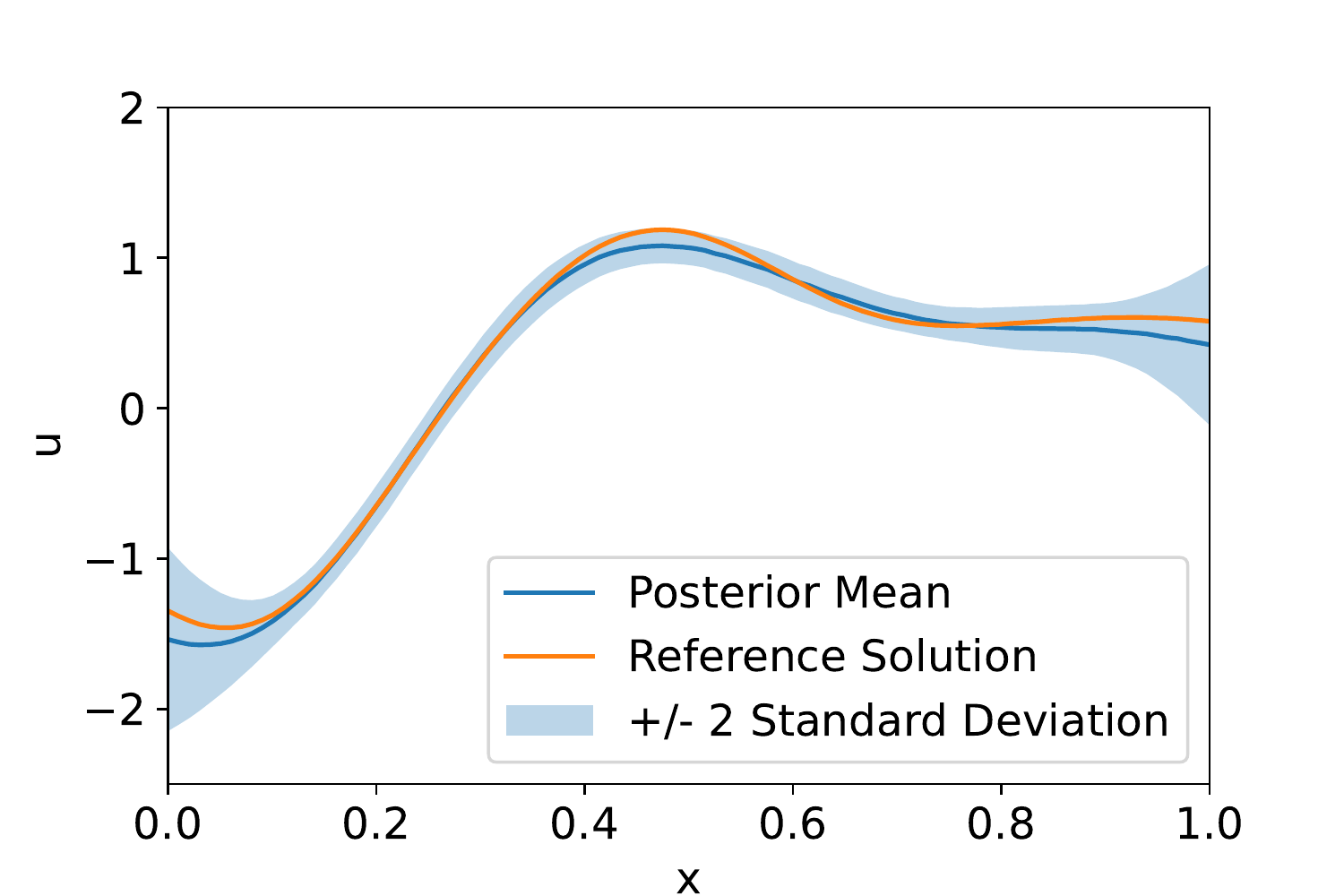}
    \includegraphics[scale=0.4]{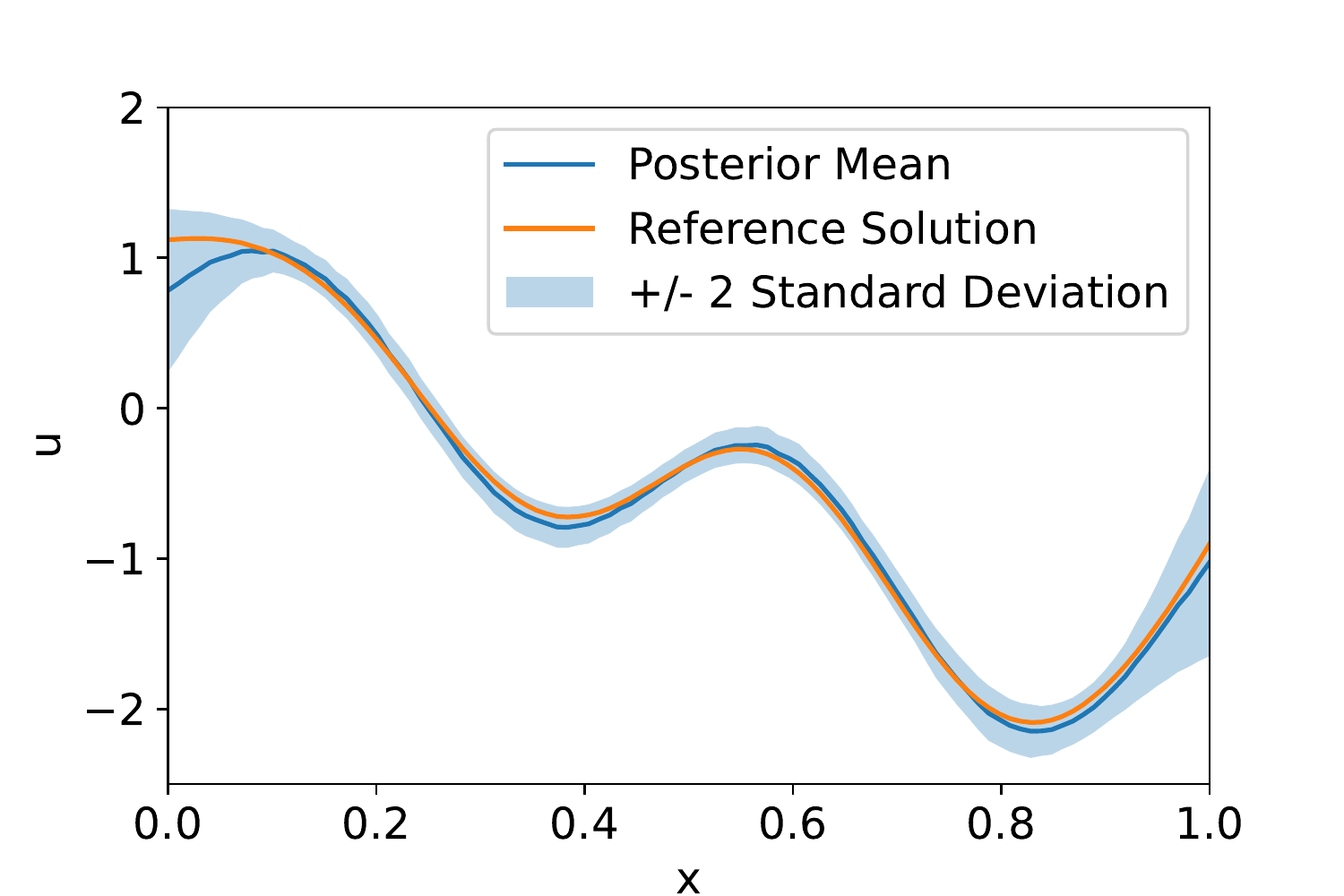}
    \caption{Bayesian Inverse Problem  for Reaction-Diffusion PDE: 100 observed data points with $\sigma^2=0.01$ and a 100-dimensional parametric input.}
    \label{fig:Inv_RD_2}
\end{figure}

\begin{figure}[h]
    \centering
    \includegraphics[scale=0.4]{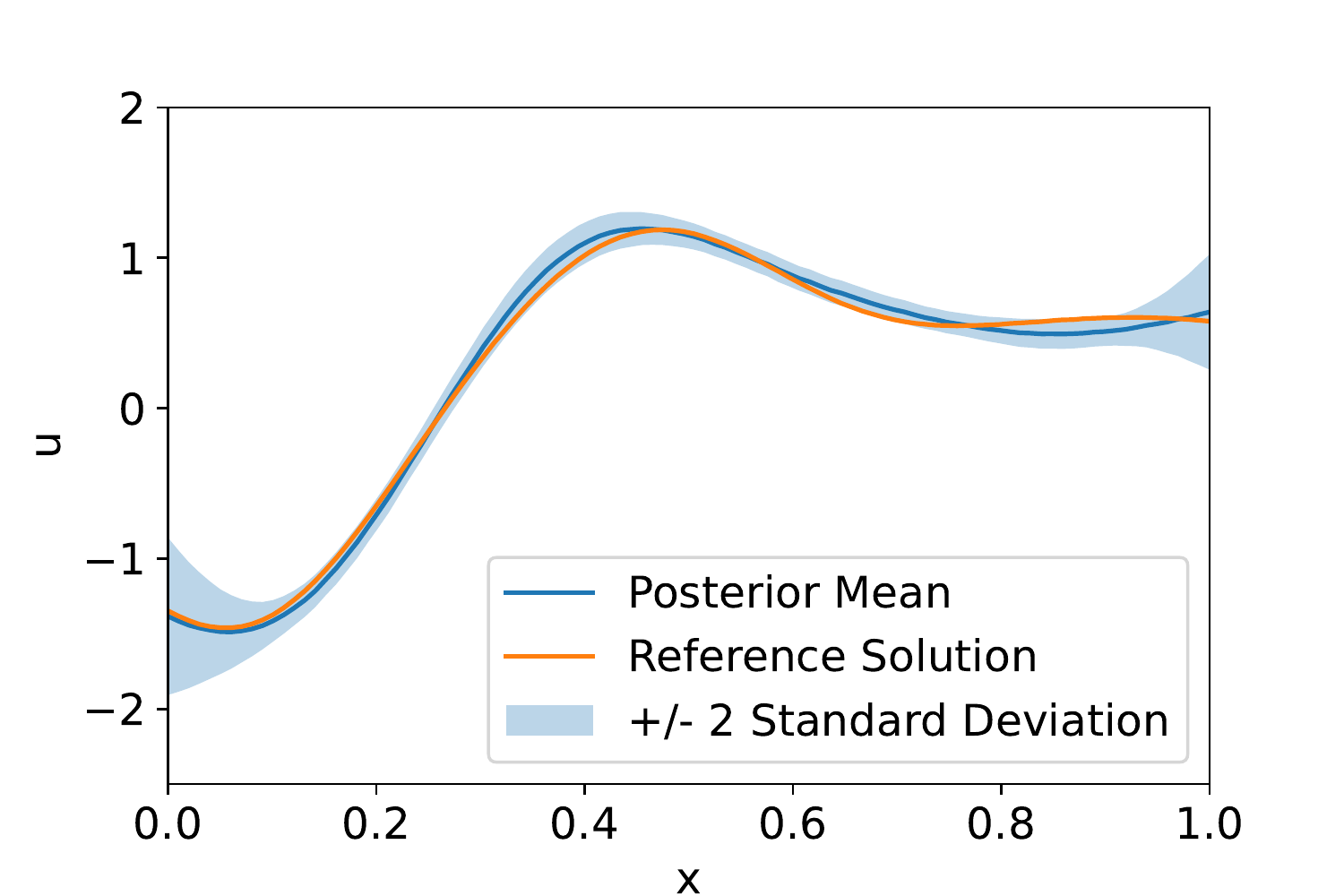}
    \includegraphics[scale=0.4]{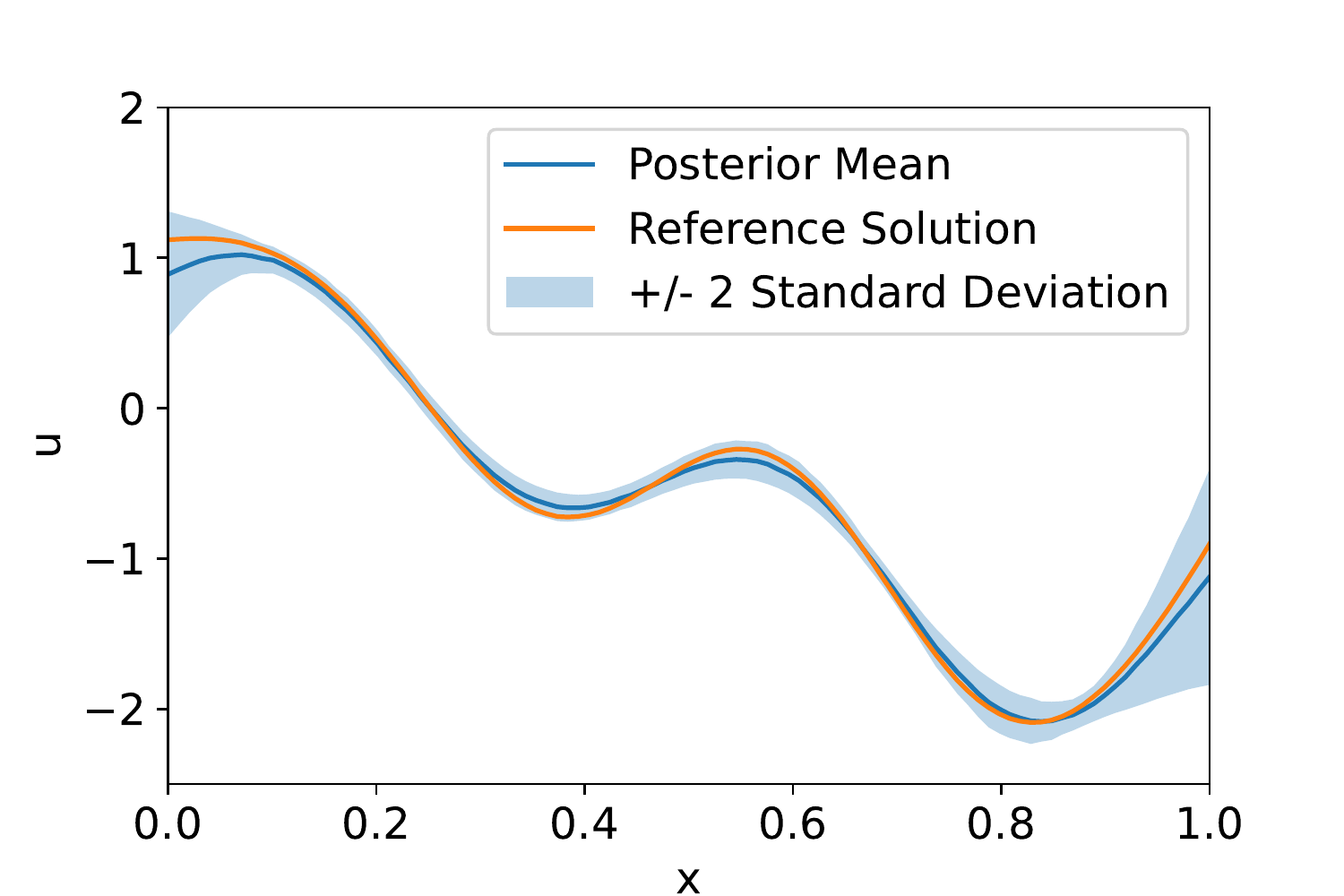}
    \caption{Bayesian Inverse Problem  for Reaction-Diffusion PDE: 25 observed data points with $\sigma^2=0.001$ and a 100-dimensional parametric input.}
    \label{fig:Inv_RD_3}
\end{figure}

We employed the  trained model of the reaction-diffusion system (with $10\%$  labeled training data), in combination with the formulation detailed in section \ref{sec:bip} for approximating the posterior.
We use a  prior $p_u(\bs{u})$ arising from the discretization of  Gaussian Process with zero mean and exponential quadratic covariance kernel with a length scale $\ell=0.2$.  For the Gaussian mixture models involved for the prior and subsequently the posterior on $\bs{b}$ we used two  components i.e. $M=2$ in Equations (\ref{eq:mixprior}), (\ref{eq:mixposterior}). The results can be seen in the following Figures. The obtained posterior encapsulates the true parameter input for all three cases.\\
In Figure \ref{fig:Inv_RD_1} we used test cases with $100$ observed solution data points for each parameter input and a noise level of $\sigma^2=0.001$ (see \refeq{eq:BIP_noise}). In Figure \ref{fig:Inv_RD_2} we increased the noise level ten-fold, to $\sigma^2=0.01$ and, as expected, so did the posterior uncertainty. 
In Figure \ref{fig:Inv_RD_3} we used $\sigma^2=0.001$ but  decreased the number of observations of the PDE-solution to $25$ points (instead of $100$). As expected, this led to an increase in posterior uncertainty.

Our method can therefore be used as a fast approach without any need for optimization and MCMC sampling to generate an approximate posterior. We note that the posterior uncertainty increases if number of observations decreases or if the observation noise  $\sigma^2$ increases.
In  Appendix \ref{sec:appendixB}, we show the excellent agreement of the approximate posterior computed with the actual one as obtained  by costly and time-consuming MCMC simulations.

\subsubsection{Flow through porous media}
\label{sec:bdarcy1}
We also solved a Bayesian inverse problem in the context of the Darcy-type PDE by using our trained model of section \ref{sec:Darcy} with added labeled training data. We computed an approximate posterior based on the algorithm presented in  section \ref{sec:bip} and compared it with the true PDE-input. For the Gaussian mixture models involved for the prior and subsequently the posterior on $\bs{b}$ we used two mixture components i.e. $M=2$ in Equations (\ref{eq:mixprior}), (\ref{eq:mixposterior}). 

Firstly, we considered permeability  fields represented with respect to $64$ known feature functions as described in section \ref{sec:Darcy}. The $64$ coefficients $c$ (\refeq{eq:features}) represented the sought PDE-inputs and a uniform prior in $[0,1]^{64}$ was employed.
The results in terms of the permeability field $u$ can be seen in the following Figures. The obtained posterior is  in good agreement with the ground truth, e.g. the PDE-input field used to generate the data with the PDE-solver.\\

\begin{figure*}
\centering
    \includegraphics[scale=0.38]{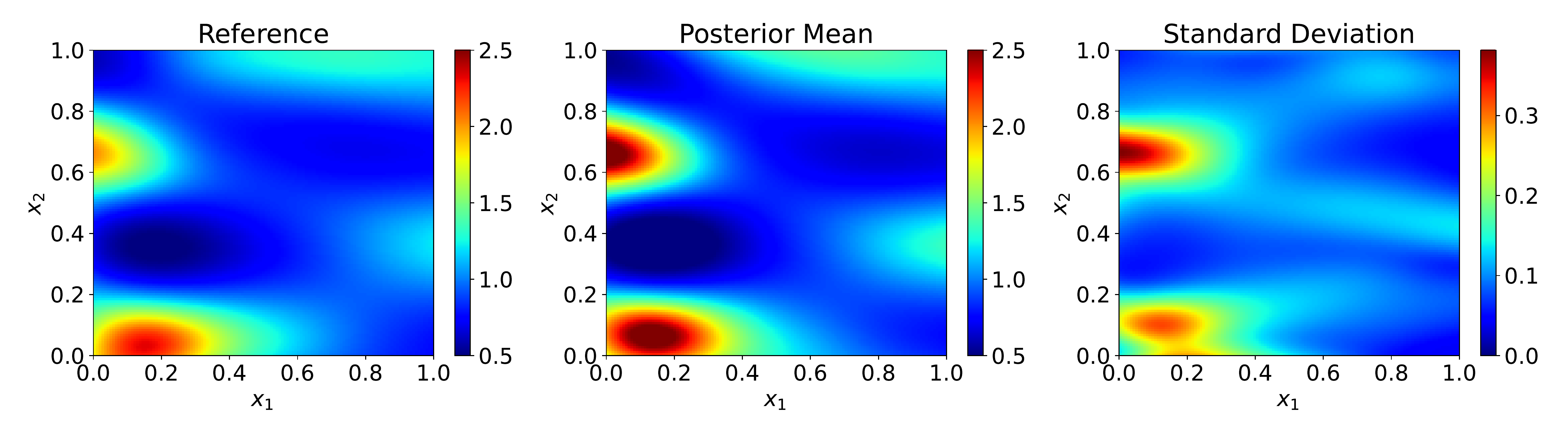}
    \includegraphics[scale=0.38]{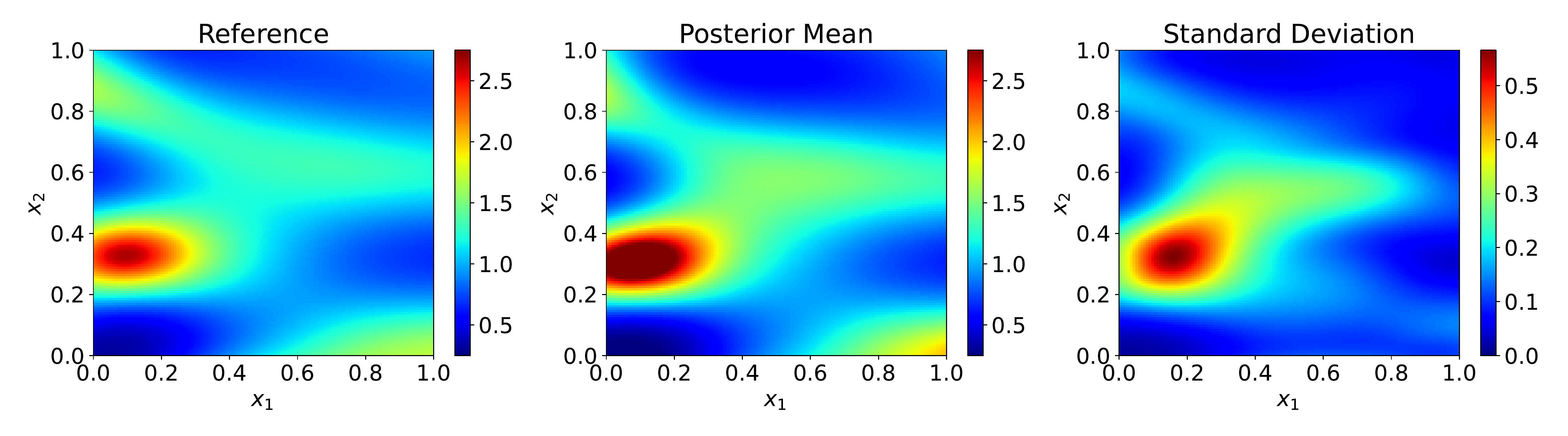}
    \caption{Bayesian Inverse Problem for Darcy-type PDE: $3844$ observed data points with $\sigma^2=0.001$ and a 64-dimensional parametric input representing feature coefficients.}
    \label{fig:Darcy_INV_f_1}
\end{figure*}

\begin{figure*}
\centering
    \includegraphics[scale=0.38]{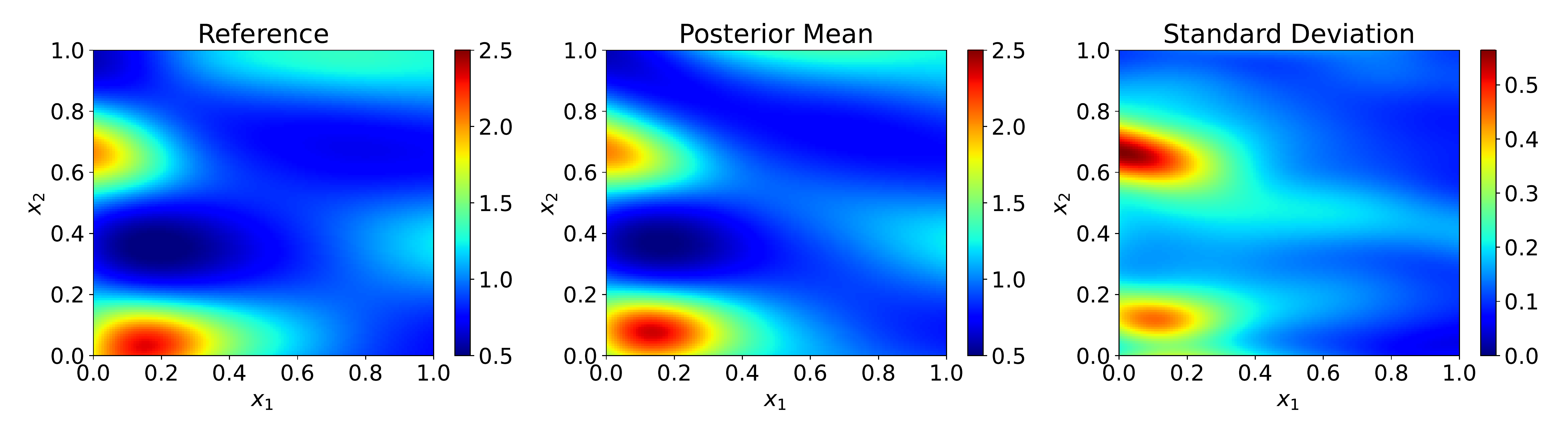}
    \includegraphics[scale=0.38]{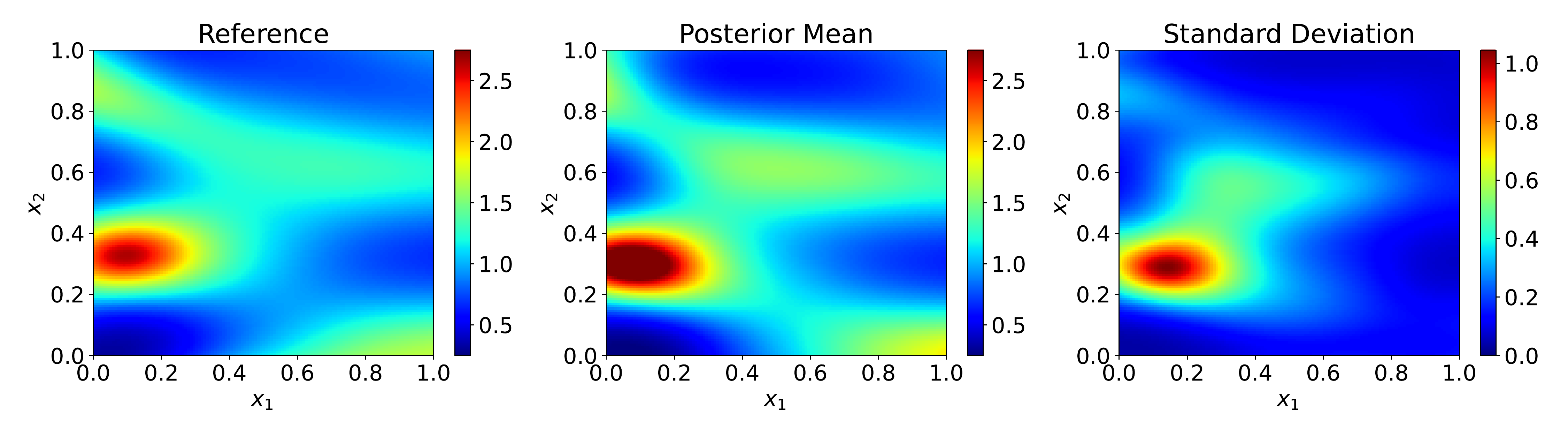}
    \caption{Bayesian Inverse Problem for Darcy-type PDE: 3844 observed data points with $\sigma^2=0.01$ and a 64-dimensional parametric input representing feature coefficients.}
    \label{fig:Darcy_INV_f_2}
\end{figure*}

\begin{figure*}
\centering
    \includegraphics[scale=0.38]{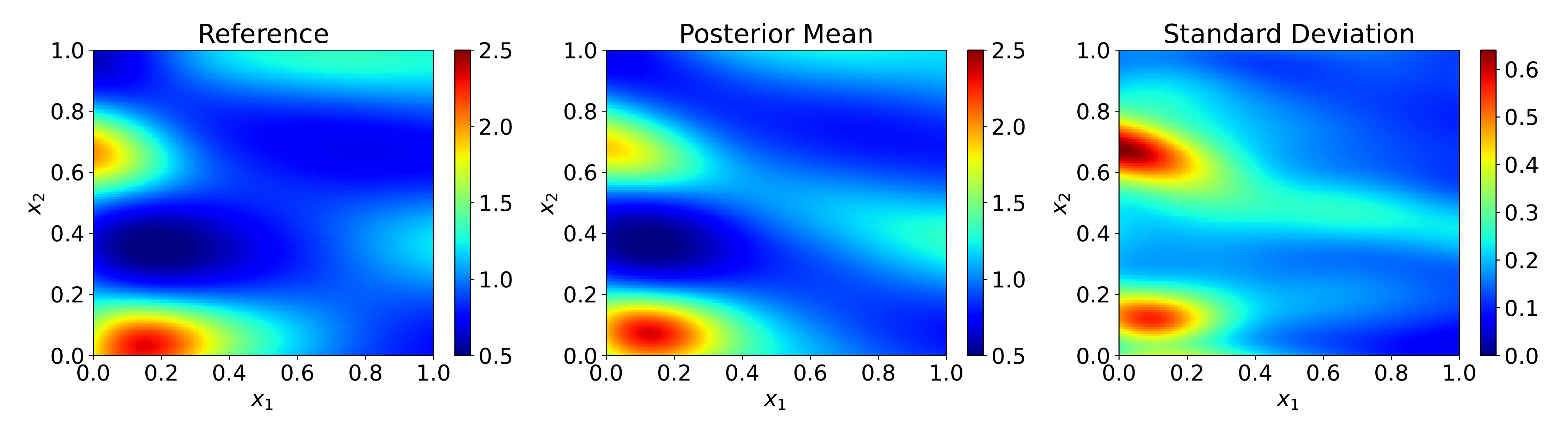}
    \includegraphics[scale=0.38]{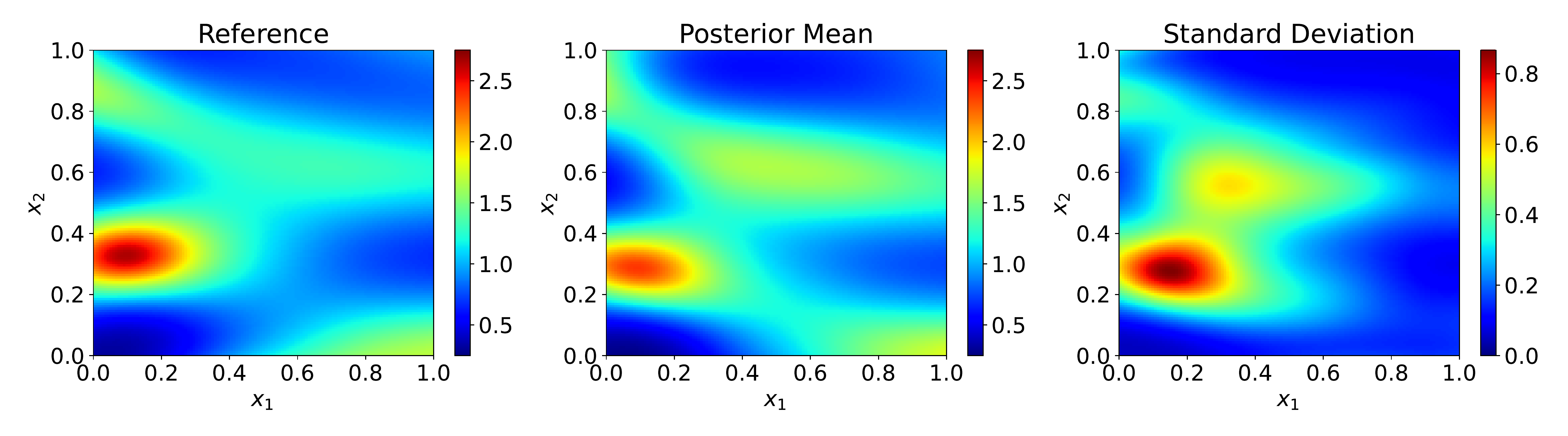}
    \caption{Bayesian Inverse Problem for Darcy-type PDE: 1922 observed data points with $\sigma^2=0.01$ and a 64-dimensional parameter input representing feature coefficients.}
    \label{fig:Darcy_INV_f_3}
\end{figure*}

In particular, in Figure \ref{fig:Darcy_INV_f_1} we assumed that $3844$ observations of the PDE-output were available, on a $62 \times 62$ regular grid. The data that was synthetically generated was contaminated with Gaussian noise with $\sigma^2=0.001$ (see \refeq{eq:BIP_noise}). In Figure \ref{fig:Darcy_INV_f_2} we increased the noise level and subsequently the posterior uncertainty was slightly higher but the posterior mean  is still close to the ground truth. In Figure \ref{fig:Darcy_INV_f_3} we  used $\sigma^2=0.01$ but decreased the number of observations by $50\%$ to $1922$. As expected, the posterior uncertainty  increased again but still encapsulated the ground truth.

\begin{figure*}
\centering
    \includegraphics[scale=0.38]{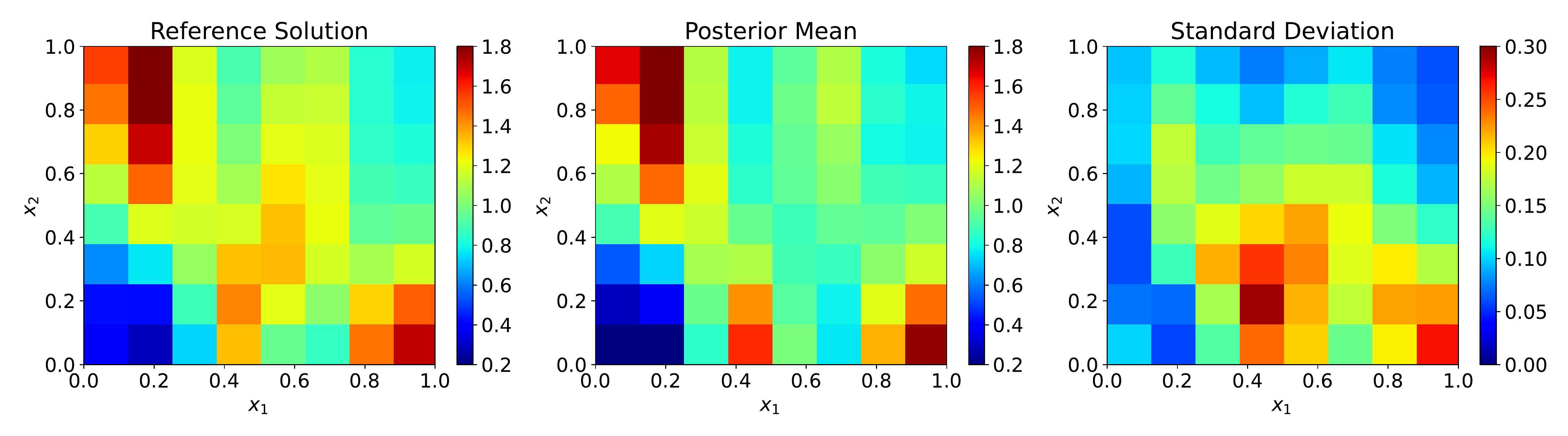}
    \includegraphics[scale=0.38]{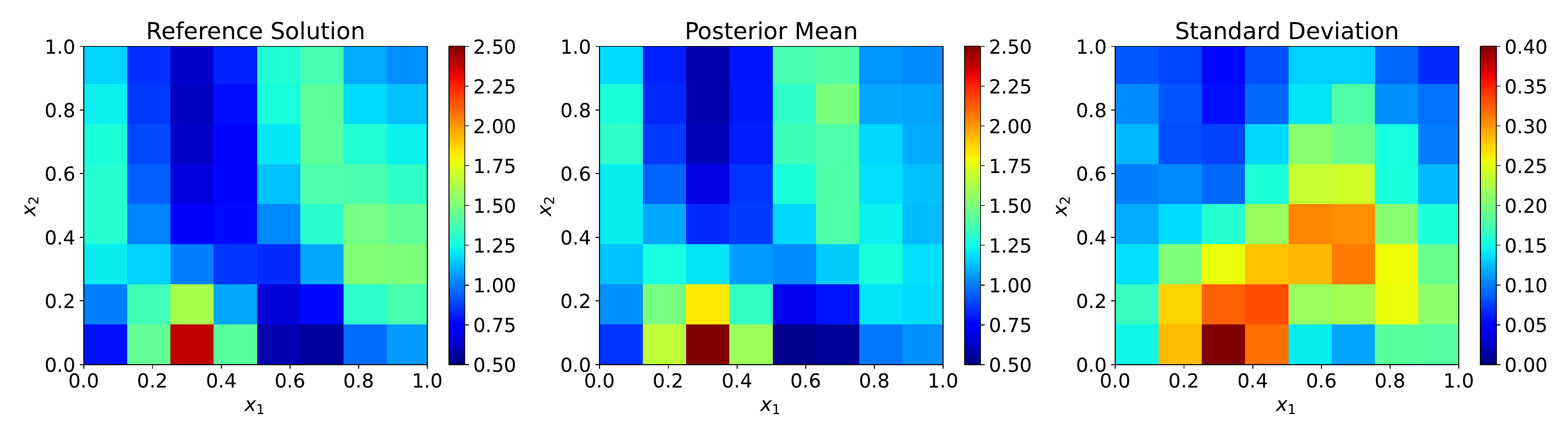}
    \caption{Bayesian Inverse Problem for Darcy-type PDE: 3844 observations with $\sigma^2=0.01$ and a 64-dimensional, discretized permeability field.}
    \label{fig:Darcy_INV_CG_1}
\end{figure*}

\begin{figure*}
\centering
    \includegraphics[scale=0.38]{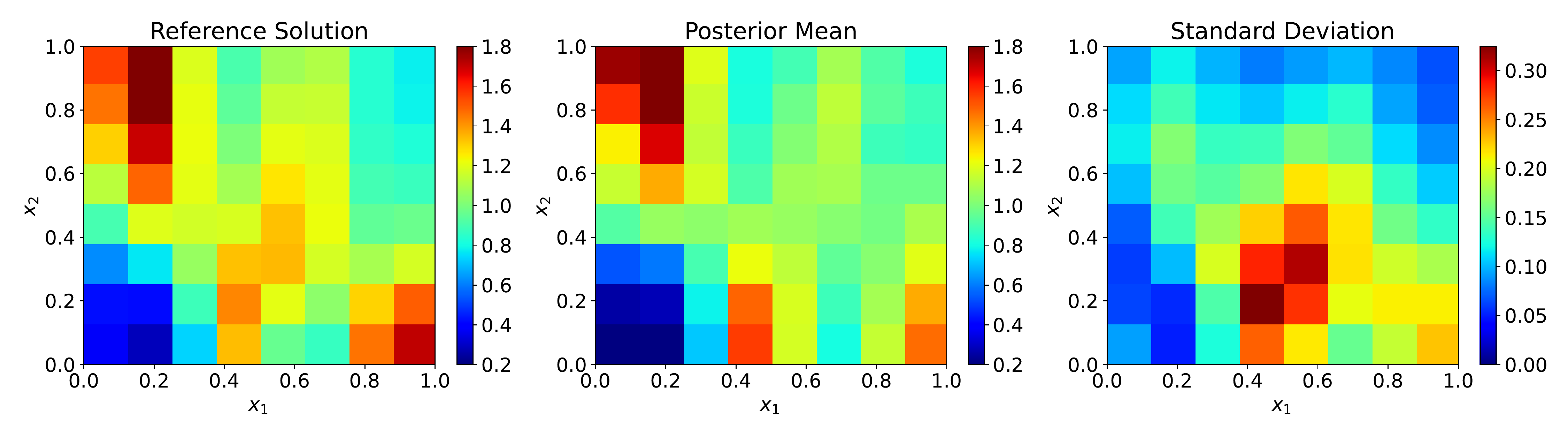}
    \includegraphics[scale=0.38]{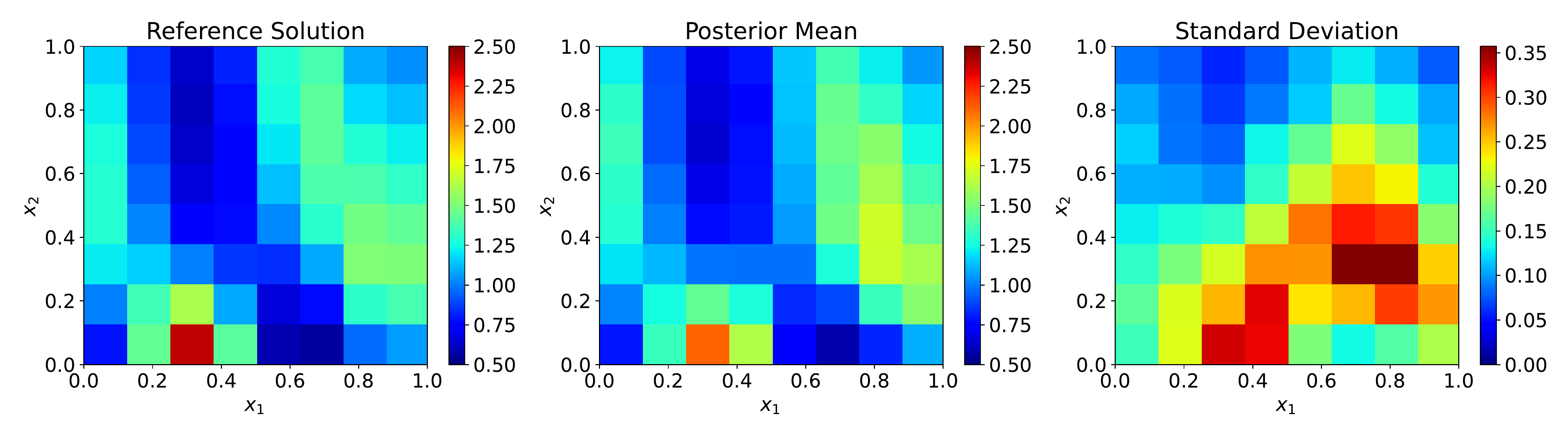}
    \caption{Bayesian Inverse Problem for Darcy-type PDE: 3844 observations with $\sigma^2=0.001$ and a 64-dimensional discretized permeability field.}
    \label{fig:Darcy_INV_CG_2}
\end{figure*}

\begin{figure*}
\centering
    \includegraphics[scale=0.38]{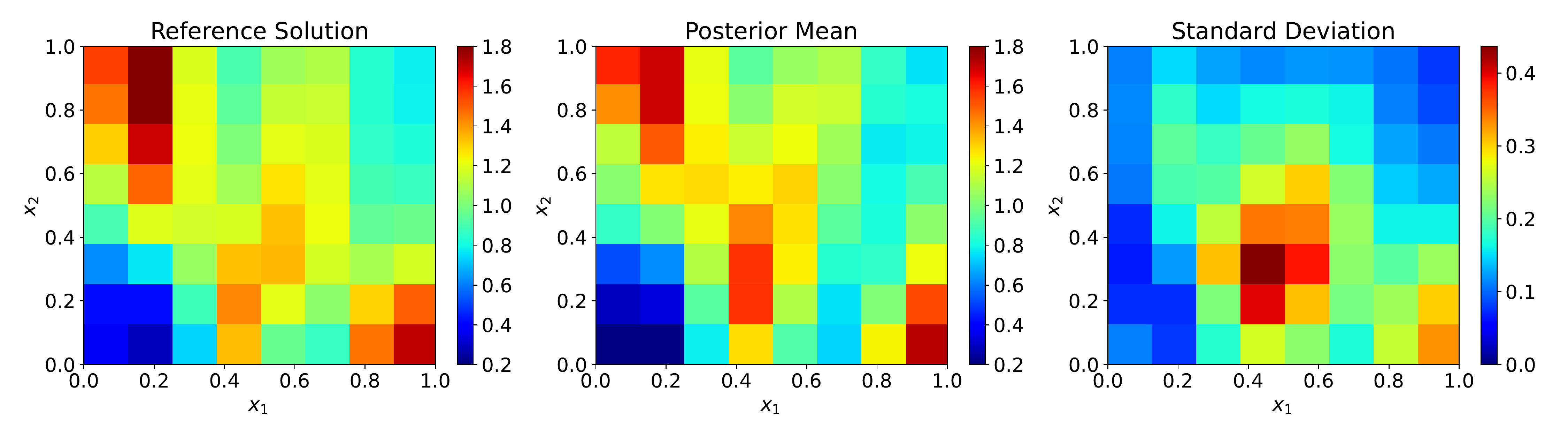}
    \includegraphics[scale=0.38]{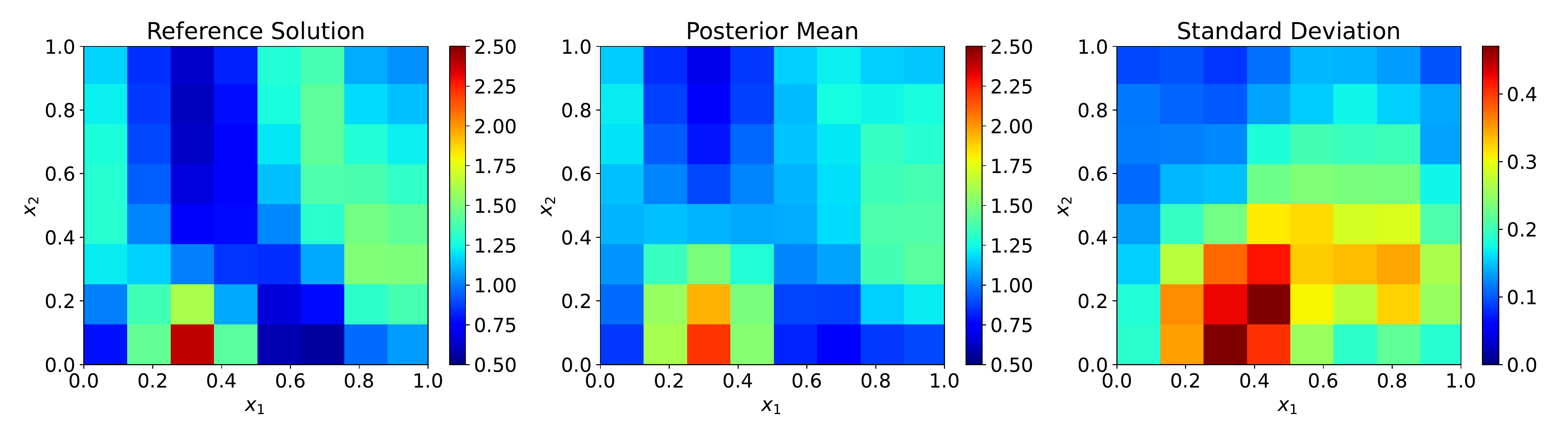}
    \caption{Bayesian Inverse Problem for Darcy-type PDE: 1922 observations with $\sigma^2=0.01$ and a 64-dimensional discretized permeability field.}
    \label{fig:Darcy_INV_CG_3}
\end{figure*}

Finally, we considered the case where the PDE-input is represented on a regular $8\times 8$ grid as in section \ref{sec:Darcy_CG}. The discretized GP described therein was used as the prior. 
In Figure \ref{fig:Darcy_INV_CG_1} we compare the ground truth with the posterior mean and standard deviation as obtained from $3844$ observations on a $62 \times 62$  regular grid and for a noise level of $\sigma^2=0.01$ (see \refeq{eq:BIP_noise}). In Figure \ref{fig:Darcy_INV_CG_2} we used lower noise with $\sigma^2=0.001$  level and, as expected,  the posterior uncertainty was  lower and the posterior mean was closer to the ground truth. In Figure \ref{fig:Darcy_INV_CG_3} we again choose the previous noise level but decreased the number of observations by half, to $1922$. As expected, the posterior uncertainty  increased but still encapsulated the ground truth.

\section{Conclusions}
\label{sec:con}
We introduced an invertible DeepONet architecture for constructing data-driven surrogates  of PDEs with parametric inputs. 
The use of the RealNVP architecture in the branch-network  enables one to obtain simultaneously accurate approximations of both the forward and the inverse map (i.e. from PDE-solution to PDE-input).
The latter is particularly useful for deterministic and stochastic (Bayesian), PDE-based, inverse problems for which accurate solutions can be readily obtained once the proposed DeepONet has been trained offline.
The training framework can make use of expensive, labeled data (i.e. PDE input-output pairs) as well as inexpensive, unlabeled  data (i.e. only PDE-inputs) by incorporating residuals of the governing PDE and its boundary/initial conditions into the loss function. The use of labeled data was generally shown to improve predictive accuracy and especially in terms of the inverse map which is something that warrants further investigation.

In the case of Bayesian formulations in particular, we showed that the availability of the inverse map can lead to highly-efficient approximations of the sought posterior without the need of additional PDE solves and without any cumbersome sampling (e.g. due to MCMC, SMC) or iterations (e.g. due to SVI).

The performance of the proposed strategy was demonstrated on several PDEs with  modest- to high-dimensional parametric inputs and its  efficiency  was assessed in terms of the amounts of labeled vs unlabeled data. Furthermore, the approximate posterior obtained was in  very good agreement with  the exact posterior obtained with the reference solver and MCMC. The accuracy persisted for various levels of noise in the data as well as when changing the number of available observations.  
We note finally that  unbiased estimates with respect to the exact posterior could be readily obtained with  Importance Sampling and by using the approximate posterior as the importance sampling density. This  would nevertheless imply additional PDE solves  which we would expect to be modest in number given the accuracy of the approximation i.e. the proximity of the Importance Sampling density with the actual posterior. 

\appendix 

\section{Influence of the amount of data}
\label{sec:appendixA}
\begin{figure*}[h]
    \centering
    \includegraphics[scale=0.5]{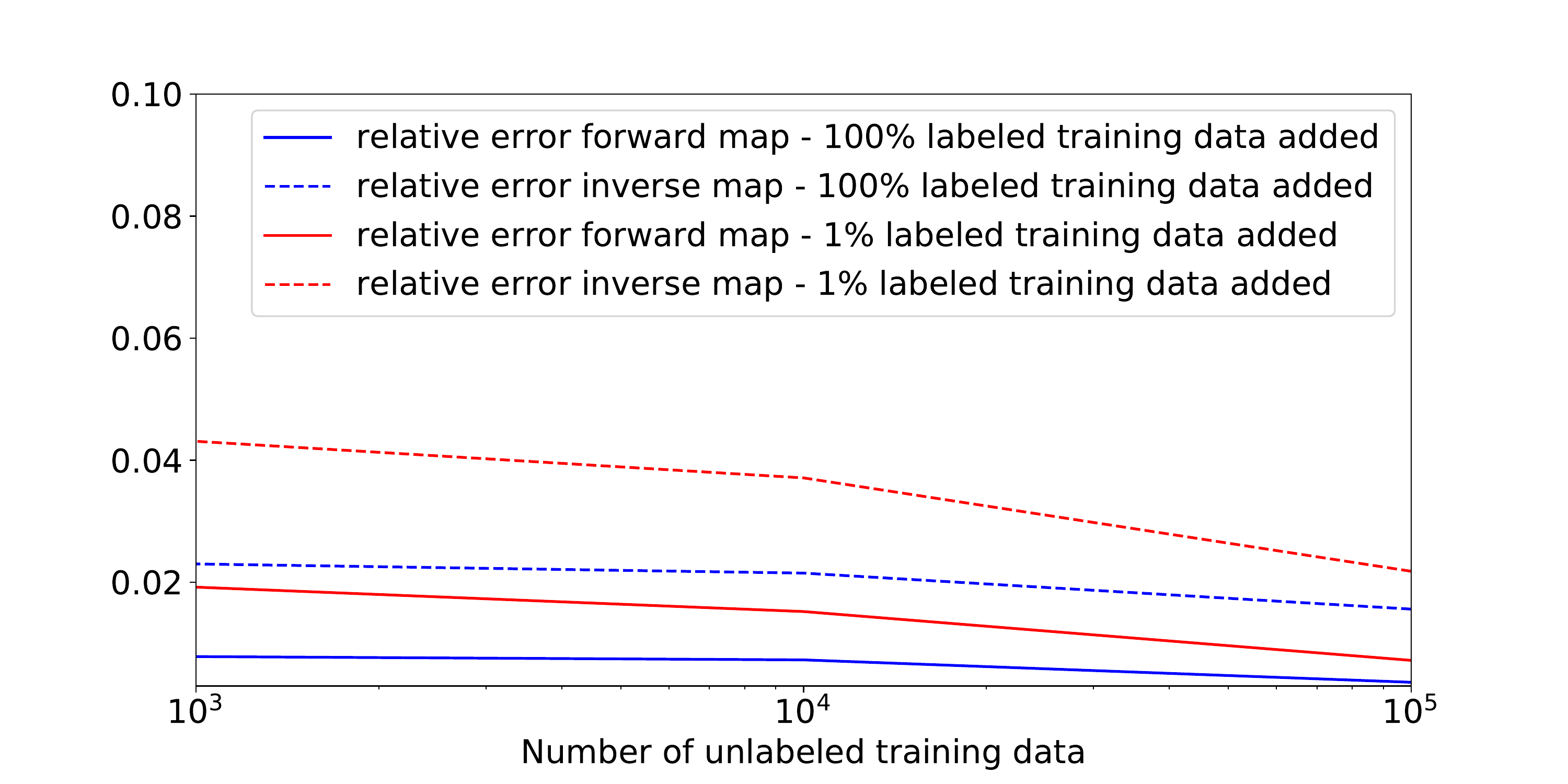}
    \caption{Relative errors on test data for the forward and inverse map depending on the amount of labeled and unlabeled training data}
    \label{fig:data}
\end{figure*}

This section contains additional results as obtained  for the antiderivative example and for different amounts of training data. We chose exactly the same settings as described in Section \ref{sec:antiderivative} and varied only the amount of labeled and unlabeled training data. In Figure \ref{fig:data} we plot the relative error in the foward and inverse map  with regards to the amount of unlabeled training data. The color indicates the amount of labeled training data used, i.e.  blue curves correspond to $1\%$  labeled training data, whereas red curves correspond to $100\%$  labeled training data.

We observe that   although the relative errors decrease  with the addition of more data, the benefit  is more pronounced with the addition of labeled data. 
\section{Comparison with MCMC}
\label{sec:appendixB}
\begin{figure*}[h]
    \centering
    \includegraphics[scale=0.5]{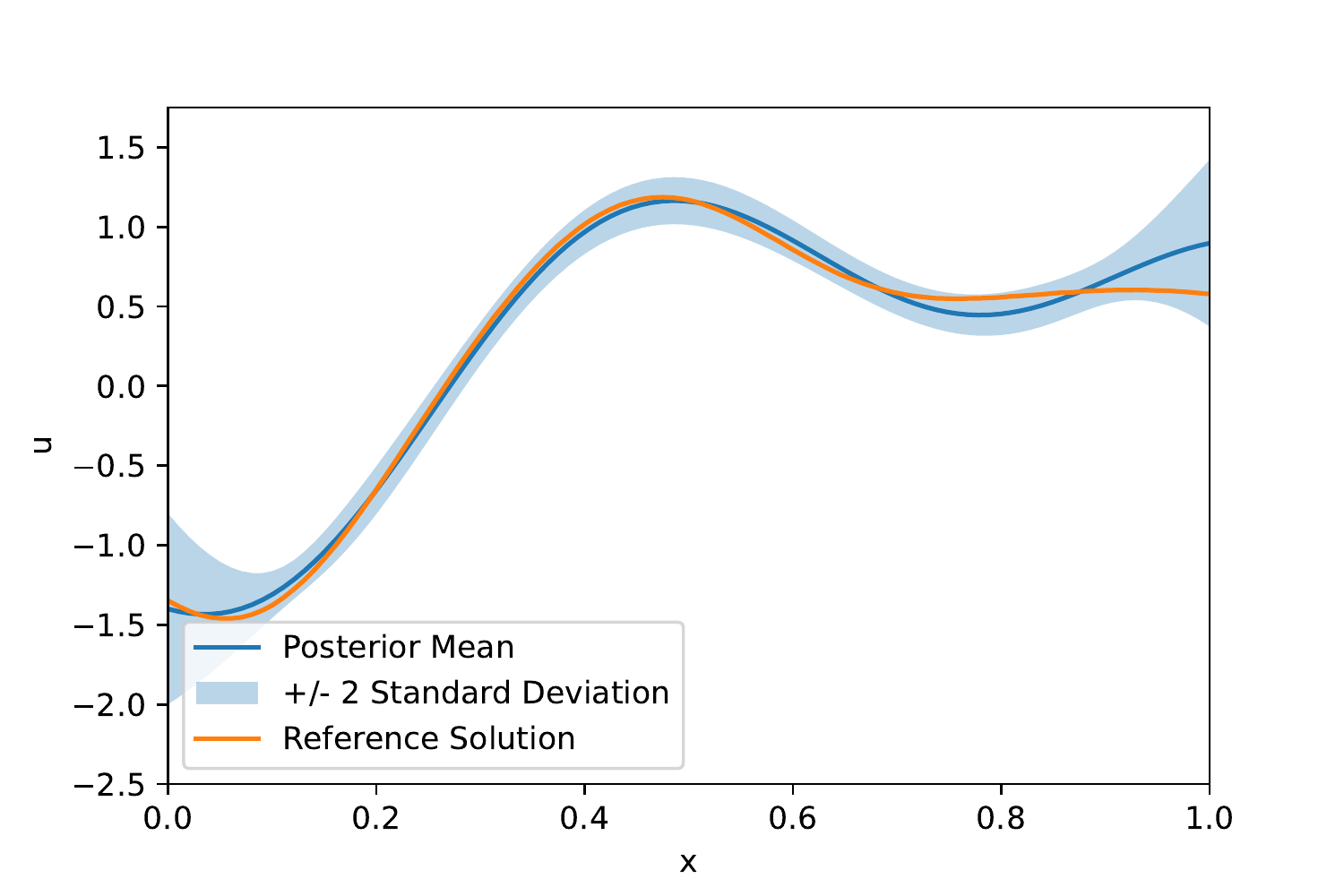}
    \includegraphics[scale=0.5]{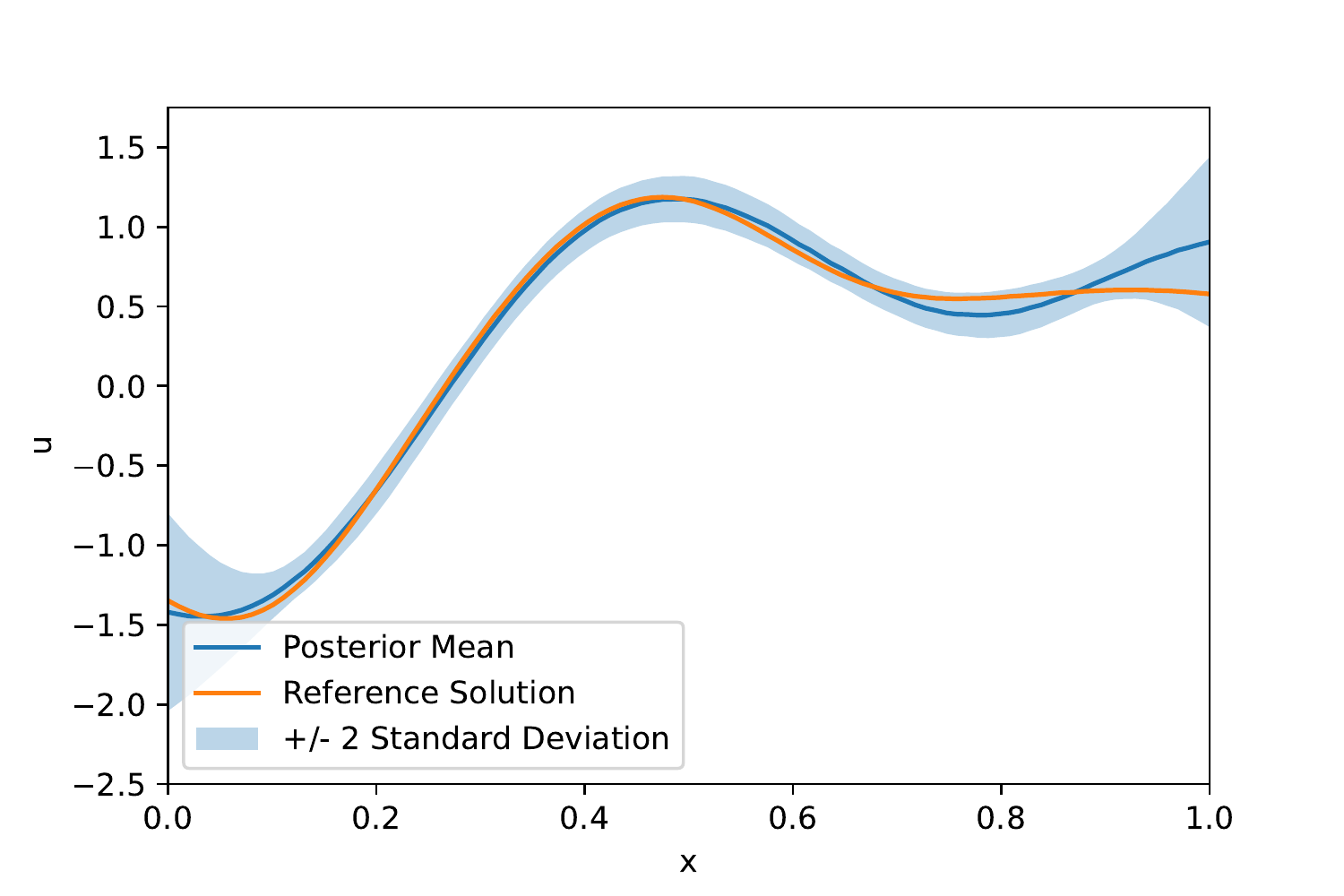}\\
    \includegraphics[scale=0.5]{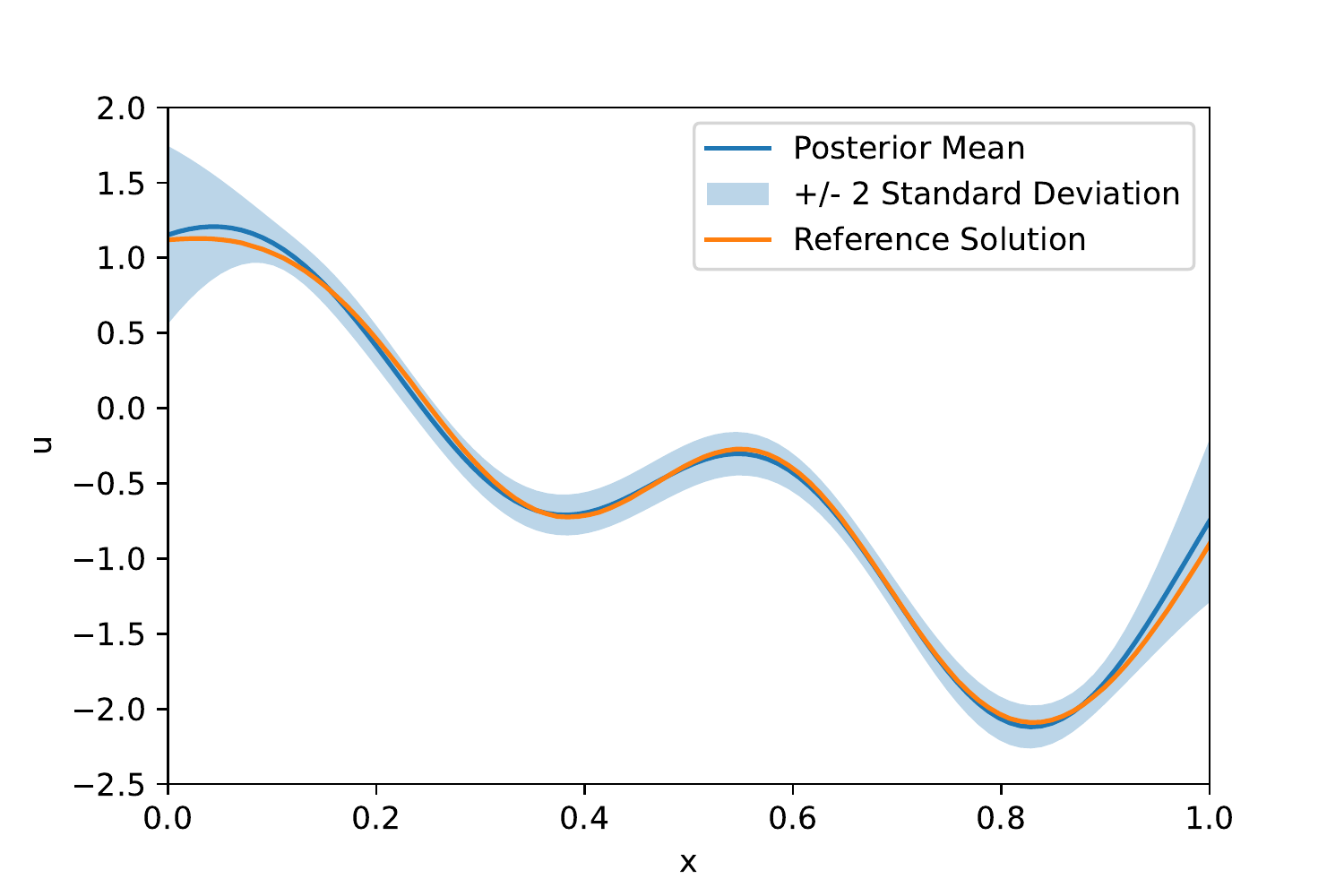}    \includegraphics[scale=0.5]{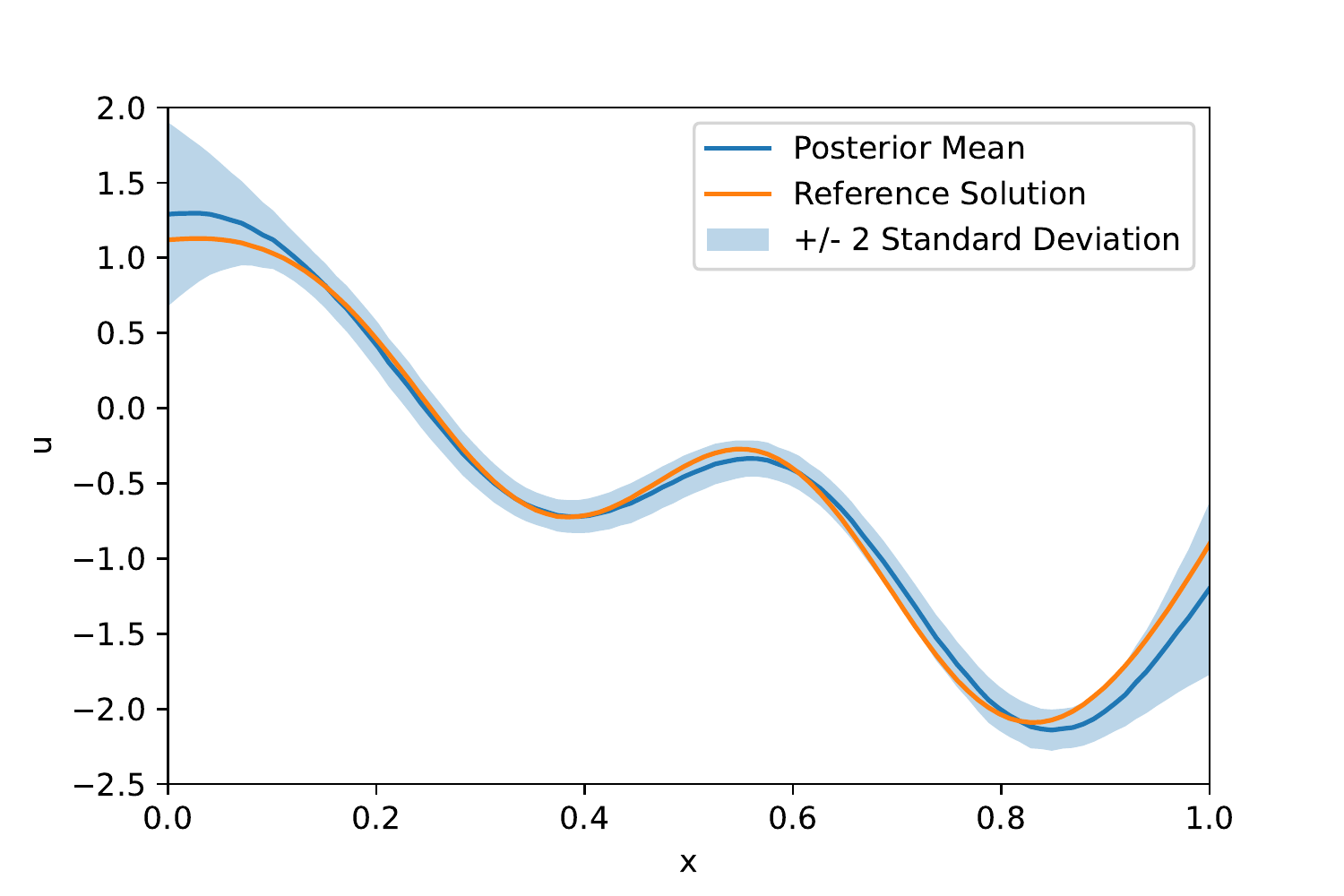}
    \caption{Bayesian Inverse Problem  for Reaction-Diffusion PDE: 100 observed data points with $\sigma^2=0.01$ and a 100-dimensional parameter input. Left: Posterior based on MCMC (NUTS), Right: Posterior obtained by our algorithm}
    \label{fig:MCMC}
\end{figure*}

In the main part of this article we already showed that the true parameter input is encapsulated by the posterior. In this section we compare the approximate  posterior computed with the reference posterior obtained by MCMC.\\
In particular, for two, randomly-chosen cases in the reaction-diffusion example, the true posterior was computed using the NUTS sampler from the Blackjax library \citep{blackjax2020github}. As is the case with all MCMC-based inference schemes, these provide the reference posterior (asymptotically). The results  shown in Figure \ref{fig:MCMC} in terms of the posterior mean $\pm$ 2 posterior standard deviations indicate excellent accuracy of the posterior approximation proposed.
While our method does not require any new forward model evaluation or model gradients, the MCMC algorithms require a forward model solve and its gradients for each sample. For the MCMC-based results displayed in total 40000 samples were generated.

\clearpage
\bibliography{sn-bibliography}


\end{document}